\documentclass[letterpaper, 10 pt, journal, twoside]{IEEEtran}
\usepackage[normalem]{ulem}
\usepackage{graphicx} 
\usepackage{epsfig} 
\usepackage{mathptmx} 
\usepackage{times} 
\usepackage{bm}
\usepackage{mathtools}
\usepackage{amsthm}
 \usepackage{nccmath}
\usepackage[font=footnotesize]{caption}
\usepackage[font=footnotesize]{subcaption}
\usepackage{float}
\usepackage{optidef}
\usepackage{amsmath} 
\usepackage{bm}
\usepackage{mathtools}
 \usepackage{nccmath}
\usepackage{amssymb}  
\usepackage[ruled,vlined,linesnumbered]{algorithm2e}
\DeclareMathOperator*{\argmax}{argmax}

\usepackage{color}
\newtheorem{remark}{Remark}
\newtheorem{theorem}{Theorem}
\newtheorem{proposition}{Proposition}
\newtheorem{lemma}{Lemma}
\newtheorem{definition}{Definition}

\usepackage{tikz}
\usetikzlibrary{shapes,arrows,positioning,fit,hobby,backgrounds,calc}
\renewcommand{\natural}{{\mathbb{N}}}

\newcommand{\integernonnegative}{\ensuremath{\mathbb{Z}}_{\ge 0}}
\newcommand{\real}{\ensuremath{\mathbb{R}}}
\newcommand{\realpositive}{\ensuremath{\mathbb{R}}_{>0}}
\newcommand{\realnonnegative}{\ensuremath{\mathbb{R}}_{\ge 0}}

\newcommand{\until}[1]{\{1,\dots, #1\}}
\newcommand{\subscr}[2]{#1_{\textup{#2}}}
\newcommand{\supscr}[2]{#1^{\textup{#2}}}

\newcommand\oprocendsymbol{\hbox{$\bullet$}}
\newcommand\oprocend{\relax\ifmmode\else\unskip\hfill\fi\oprocendsymbol}

\renewcommand{\baselinestretch}{0.99}

\begin{document}
\bstctlcite{IEEEexample:BSTcontrol}
%
\title{Robotic Exploration using \\ Generalized Behavioral Entropy}
%
%
%

\author{Aamodh Suresh, Carlos Nieto-Granda, Sonia Mart{\'i}nez 
\thanks{Manuscript received: Feb, 15, 2024; Revised May, 22, 2024; Accepted July, 1, 2024.}
\thanks{This paper was recommended for publication by Editor Hanna Kurniawati upon evaluation of the Associate Editor and Reviewers' comments.
This work was supported by U.S. Army Research Laboratory Grant W911NF-23-2-0009.}
\thanks{Aamodh Suresh and Carlos Nieto-Granda are with the U.S. Army Research Laboratory (ARL), Adelphi, Maryland, aamodh@gmail.com, carlos.p.nieto2.civ@army.mil. Sonia Mart{\'i}nez is with the Department of Mechanical Engineering, UC San Diego, California, soniamd@ucsd.edu.}
}
%
%

\markboth{IEEE Robotics and Automation Letters. Preprint Version. Accepted July, 2024}
{Suresh \MakeLowercase{\textit{et al.}}: Behavioral Entropy}

%



\maketitle

\begin{abstract}
This work presents and evaluates a novel strategy for robotic exploration that leverages human models of uncertainty perception. 
To do this, we introduce a measure of uncertainty that we term ``Behavioral entropy'', which builds on Prelec's probability weighting from Behavioral Economics. 
We show that the new operator is an admissible generalized entropy, analyze its theoretical properties and compare it with other common formulations such as Shannon's and Renyi's. 
In particular, we discuss how the new formulation is more expressive in the sense of measures of sensitivity and perceptiveness to uncertainty introduced here.  
Then we use Behavioral entropy to define a new type of utility function that can guide a frontier-based environment exploration process. 
The approach's benefits are illustrated and compared in a Proof-of-Concept and ROS-Unity simulation environment with a Clearpath Warthog robot. We show that the robot equipped with Behavioral entropy explores faster than Shannon and Renyi entropies.
\end{abstract}


%
\IEEEpeerreviewmaketitle

\section{INTRODUCTION}
\label{sec:introduction}

\tikzstyle{block} = [rectangle, draw, fill=blue!20, 
text width=5em, text centered, rounded corners, minimum height=2.5em]
\tikzstyle{bigblock} = [rectangle, draw, fill=blue!20, 
text width=7.5em, text centered, rounded corners, minimum height=2.5em]    
\tikzstyle{line} = [draw, -latex']
\tikzstyle{cloud} = [draw, ellipse,fill=red!20, node distance=3cm,
minimum height=2em]

\IEEEPARstart{A}{utonomous} robotic exploration is critical as  
it alleviates the inaccessibility and dangers of remote, unsafe, and risky environments. 
Depending on the environment and task at hand, the robot might be expected to exhibit a variety of exploration behaviors from coarse and fast to fine and detailed. 
These tasks are invariably evaluated or supervised by humans and the exploration objective could change dynamically. Thus having an intuitive, diverse and theoretically grounded exploration framework is a necessity. 
In this work, we infuse human perception characteristics into robotic exploration AI, and investigate both theoretically and practically, the effects of this infusion.

The fundamental approach to exploration consists of reducing uncertainty by observing an area with noisy sensors. 
Uncertainty is typically quantified by entropy~\cite{shannon1948mathematical} and majority of exploration policies aim at reducing it (implicitly or explicitly). 
However, unlike the typical robotic AI designs, it is well known that humans perceive uncertainty in a fundamentally non-rational manner~\cite{DP:98,AT-DK:92,AS-AT-LR-SM:23}, especially in sensory perception and evaluating outcomes~\cite{DB-RD:12-hum-unc-quant-nat}, causing a variety of behaviors and decision making. 
Motivated by this, we first propose and characterize a novel measure of uncertainty called Behavioral entropy that incorporates human models of uncertainty perception~\cite{DP:98} that is widely used in the Behavioral Economics literature. Then, we investigate the design of new behavioral robotic exploration policies using our proposed  uncertainty measure.
\paragraph*{Related Work}
Robotic exploration broadly involves the cyclic interplay between \emph{perception} (knowledge of environment and self via SLAM~\cite{ST-WB-DF:05}), \emph{decision} (evaluating areas of interest (AOIs) from current knowledge) and \emph{action} (navigation policies to reach the selected AOI). 
Several SLAM techniques have been proposed to produce environment maps utilizing visual odometry~\cite{tian2022kimera} and LiDAR~\cite{nieto2014IJRR} that are key components for exploration. 
Similarly, there exists navigation pipelines utilizing a combination of global planning~\cite{AS-SM:21-ral,AS-CN-SM:23-purrrrt}, local planning~\cite{mppi_control:17}, optimal control~\cite{koga2021active} and reactive control~\cite{AS-SM:22-lcss} to reach an AOI safely, reliably and efficiently. We focus on the \emph{decision} aspect of identifying and evaluating potential AOIs to navigate from a given map.     
In particular, we use the popular frontier-based exploration~\cite{yamauchi1998frontier}, that has been successfully used to explore challenging environments~\cite{sun2020frontier,otsu2020supervised}. 
These strategies are driven by optimizing utility functions that measure notions of information gain or uncertainty reduction by {\color{black}visiting} these AOIs. 
Thus, the metrics employed in these utility functions are a key factor in designing exploration policies. 
The most commonly used metric is based on the Shannon information gain~\cite{yamauchi1998frontier} that uses the Shannon entropy~\cite{shannon1948mathematical} as a measure of uncertainty. 
More recently, information metrics derived from Renyi's entropy have been proposed
and shown to perform better than Shannon's information gain~\cite{carrillo2018autonomous}. 
Renyi's entropy generalizes Shannon's to a family of generalized entropies~\cite{ribeiro2021entropy}. 
However, these entropies and the corresponding metrics do not incorporate human behavioral models. 
Moreover, their tuning is non-intuitive and Renyi's entropy is not continuously differentiable w.r.t. its parameter (at $1$) and thus making tuning also difficult.
Some works try to address modelling human choice and uncertainty perception through data intensive approaches like Boltzmann machines~\cite{osogami2014restricted} and Bayesian learning~\cite{AK-JF-ML:21-unc-bay}. 
However data-based approaches require large datasets and computational effort, provide limited theoretical insights, can suffer from poor generalization and are non-intuitive to tune and implement.
Thus, we formulate a new entropy that is based on human models of uncertainty perception and which provides the most broad set of uncertainty perception tools to tackle the exploration problem, while also being smooth, theoretically grounded and intuitive to tune.  

\paragraph*{Contributions} 
In this work, we introduce a novel operator ``Behavioral entropy'' to quantify subjectively perceived uncertainty of a given probability distribution. 
Our entropy design is intuitive and empirically grounded from Behavioral Economics, as well as ``admissible'' as a generalized entropy and theoretically grounded from Information theory perspective.
We provide novel measures of ``Sensitivity'' and ``Perceptiveness'' to characterize behavioral uncertainty perception of entropic measures; which provides a basis to characterize and compare any admissible entropic measure. 
We show that Behavioral entropy is more general in the sense that it can capture the widest range of uncertainty "perceptions" w.r.t. Shannon's and Renyi's (which is made more precise later). We also show that in this regard, Renyi's entropy is more general than Shannon's.
Then, we construct a novel exploration utility function based on Behavioral entropy and combine it into a frontier-based exploration approach. 
We further analyze and show that our exploration formulation is more expressive and can capture a wider range of perceived utility than others that use standard {\color{black}Bolzmann-Gibbs-Shannon (BGS)} and Renyi entropies. 
We perform proof of concept simulations {\color{black} with ideal SLAM and robot navigation} in a custom occupancy map to evaluate our proposed method in isolation, and show that our method is faster than methods with other entropies. 
Finally, we evaluate our framework in a {\color{black} complex and realistic} Unity-ROS simulation environment used in the DARPA Subterreanean challenge~\cite{darpasubt2021}, with a Clearpath Warthog robotic system with a LiDAR and IMU based SLAM system. 
We show that the robot equipped with our behavioral entropy based method, explores the environment quicker than other methods. 
\section{Generalized Entropy and Problem Formulation}
\label{sec:problem_formulation}

We start introducing basic notations\footnote{We let $\real$ denote the set of real numbers, $\integernonnegative$ the
  set of positive integers, and $\realnonnegative$ are non-negative
  real numbers.  $\mathbf{P}(A)$ is the power set of any other set $A$. $\mathbf{B}_r^{x}$ denotes the $n$ dimensional ball of radius $r$ centered at $x \in \real^n$.  }, a concise description of generalized entropy, 
followed by our problem formulation.

\paragraph*{Generalized entropy}
\label{sec:prelim_entropy_axioms}
\label{sec:prelims}

Entropy is a measure of uncertainty associated with the probability distribution of a random variable. 
More precisely, consider the set of probability distributions $\mathcal{P}_M$ of discrete random variables of $M$ outcomes; that is, $(p_1,\dots, p_M) \in \mathcal{P}_M$, if $0 \le p_i  \le 1$, for $i = 1,\dots, M$, and $\sum_{i=1}^M p_i=1$. 
Let $H : \mathcal{P} \triangleq \cup_{M=1}^\infty \mathcal{P}_M \rightarrow \realnonnegative$ be a permutation-invariant operator. 
We say that $H$ belongs to the class of entropy functions if it satisfies the following Shannon-Kinchin axioms~\cite{JA-SB-SH:18,shannon1948mathematical}:

\begin{enumerate}
    \item[A1] \emph{Continuity:} $H$ is continuous over each $\mathcal{P}_M$. 
    \item[A2] \emph{Maximality:} $H(p) \leq H(\frac{1}{M},\dots,\frac{1}{M})$, and the equality holds if and only if
    $p_i = \frac{1}{M}$, for all $i\in \until{M}$. This axiom implies that the uniform distribution attains the highest entropic 
    measure.
    \item[A3] \emph{Expansibility:}  A sure outcome added to a probability distribution $p$ does not change its entropy. i.e.:\\
    $H(p_1,\dots,p_i,0,p_{i+1},\dots,p_M)=H(p_1, \dots,p_M)$. 
    \item[A4] \textit{Separability or Strong Additivity:} For any two $p,q \in \mathcal{P}_M$, let 
    $H(p,q)$ denote the entropy of a joint distribution of $p,q$, then 
    $H(p,q)=H(p) + H(q|p)$, where $q|p$ is the conditional distribution of $q$ given $p$. 
\end{enumerate}
Shannon proved~\cite{shannon1948mathematical} that an entropic function satisfying all the four axioms is necessarily a scaled BGS entropy~\eqref{eqn:entropies_shannon}.
 We focus on generalized entropies defined as follows. 
\begin{definition}[Generalized entropy]
    \label{def:generalized_entropy}
    An entropic function $H$ is an \textit{admissible generalized entropy}, if it satisfies the first three entropy axioms A1-A3.    
\end{definition}
 BGS (Shannon) entropy $\supscr{H}{S}$ and Renyi entropy $\supscr{H}{R}$ ( generalises many other entropies like Tsallis entropy~\cite{ribeiro2021entropy}) are well known admissible generalized entropic functions, defined as follows:

\begin{subequations} \label{eqn:entropies}
\centering
    \begin{align}
       \supscr{H}{S}(p) =& -k\sum_{i=1}^M p_i \log(p_i)  ,\ k \in \realpositive,\ \label{eqn:entropies_shannon}  \\
       \supscr{H}{R}(p) =& \frac{1}{1-\gamma}\log \left( \sum_{i=1}^M p_i^\gamma \right) ,\ \gamma \in \realpositive,\ \gamma \neq 1, \label{eqn:entropies_renyi} 
    \end{align}
\end{subequations}

\paragraph*{Robotic Exploration Setting}
\label{sec:exploration_intro}
Our application of interest is robotic exploration. To this end, we represent the environment where the robot is deployed as a compact set $\mathcal{X} \subset \real^2$. 
The discretization of $\mathcal{X}$ using square grid cells results into the set of grid elements $\mathcal{D}$ and an associated occupancy map $\mathcal{M}$ representing the likely location of obstacles in the environment. 
More precisely, consider an occupancy function $\subscr{f}{occ} : \mathcal{D} \rightarrow [0,1]$, which assigns probability of occupancy of each cell in $\mathcal{D}$. The occupancy map $\mathcal{M}$ is the discrete field of $\subscr{f}{occ}$ over $\mathcal{D}$. 
We partition $\mathcal{D}$ and, subsequently, $\mathcal{M}$ into the unknown space $ \supscr{\mathcal{D}}{UK} \triangleq \{x \in \mathcal{D}: \subscr{f}{occ}(x) \triangleq 0.5\}$ (resp.~$\supscr{\mathcal{M}}{UK}$), the uncertain space 
$ \supscr{\mathcal{D}}{U} \triangleq \{x \in \mathcal{D}: \subscr{f}{occ}(x) \in (0,0.5) \cup (0.5,1)\}$ (resp.~$\supscr{\mathcal{M}}{U}$), and the known space $ \supscr{\mathcal{D}}{K} \triangleq \{x \in \mathcal{D}: \subscr{f}{occ}(x) \in \{0,1\}\}$ (resp.~$\supscr{\mathcal{M}}{K}$). 
The process of ``exploration'' aims to reduce the volume of unknown space $\supscr{\mathcal{D}}{U}$ and increasing the volume of known space $ \supscr{\mathcal{D}}{K}$. In an ideal setting, exploration is complete when $\supscr{\mathcal{D}}{K} =\mathcal{D}$. However, factors like sensing noise and obstacles might hinder uncertainty reduction and could make unknown areas unobservable and remain uncertain. 

The robot uses LiDAR 
to perceive its immediate environment, IMU and odometry to approximate its internal state. We use the popular SLAM~\cite{ST-WB-DF:05} method called ``Omnimapper''~\cite{nieto2014IJRR}
to simultaneously estimate the robot's pose $x \in SE(2)$ and estimated occupancy map $\mathcal{M}$. 
The LiDAR provides sensor measurements of an area $\mathcal{A} \subseteq \mathcal{D}$ around the robot and the SLAM algorithm updates $\mathcal{A}$ in $\mathcal{M}$ using physical models of sensor.

The frontiers $F\triangleq\{f_1,\dots,f_N\}$ where $f_i \in SE(2)$ are identified from the current occupancy map $\mathcal{M}$, and contain locations to explore next.  
We employ a utility function $u: SE(2) \times SE(2) \times \mathbf{P}(\mathcal{M}) \rightarrow \realnonnegative$  to quantify the reward of the robot being at a pose $x$ and observing an area $\mathcal{A}\subset \mathcal{M}$ around a frontier $f_i$; $u(x,f_i,A) \in \realnonnegative$. In this work, the rewards of exploring an area $A$ are given in terms of an information gain. 
After identifying the best frontier to go next, the navigation manager in the robot plans global paths and local control actions to reach the selected frontier location. 

\begin{figure}[h]
	\centering
	\resizebox{0.95\linewidth}{!}{%
		
		\begin{tikzpicture}[node distance = 1.6cm, auto]
			\node [block] (slam) {SLAM};
            \node [block, above right=-0.25cm and 0.5cm of slam,align=center] (frontier) {Frontier Extraction};
            \node [block, below right=-0.25cm and 0.5cm of slam,align=center] (perceived) {Perceived Occupancy};
            \node [block, right =0.5cm of frontier] (cluster) {Frontier Clustering};
            \node [block, right =0.5cm of perceived] (behentropy) {Behavioral Entropy};
            \node [block, below right=-0.25cm and 0.5cm of cluster,align=center] (infogain) {Behavioral Info Gain};
            \node [block, below =1.5cm of infogain] (utility) {Behavioral Utility};
            \node [block, left =0.5cm of utility] (explorer) {Behavioral Explorer};
            \node [block, left =0.5cm of explorer] (navmanager) {Navigation Manager};
            \node [block, left =0.5cm of navmanager] (robot) {Robot Controller};
            \begin{pgfonlayer}{background}
            \draw[blue,fill=red,opacity=0.2](perceived.north west) to
            [closed,curve through={ 
            ($(perceived.north east)!0.5!(cluster.west)$)
            .. (cluster.north west) 
            .. (cluster.north east) 
            .. (infogain.north east) 
            .. (infogain.south east)
            .. (utility.south east)
            .. (utility.south west)
            .. (explorer.south east)
            .. (explorer.south west)
            .. (explorer.north west)
            .. (perceived.south west)}]
            (perceived.north west);
            \end{pgfonlayer}

			
			
			\draw[->] (slam.east) -| 
                 ([xshift=-2.5mm]frontier.west) -- node[below=0.4cm]{$\mathcal{M}$} (frontier.west);
            \draw[->] (slam.east) -| 
                 ([xshift=-2.5mm]perceived.west) -- (perceived.west);
            \draw[dashed,->] (slam.north) -- node[midway,right]{$\mathcal{L}$}([yshift=0.5cm]slam.north);
            \draw[dashed,->] (slam.south) -- node[midway,left]{$x$}([yshift=-0.5cm]slam.south);
            \draw[->] (frontier.east) -- node[midway,above]{$f$} (cluster.west);
            \draw[->] (perceived.east) -- node[midway,above]{$\mathcal{W}$} (behentropy.west);
            \draw[->] (cluster.east) -| 
                 ([xshift=-2.5mm]infogain.west) -- node[above=0.75cm]{$F$} (infogain.west);
            \draw[->] (behentropy.east) -| 
                 ([xshift=-2.5mm]infogain.west) -- node[below=0.5cm]{$\supscr{H}{B}$} (infogain.west);
            \draw[->] (infogain.south) -- node[midway,right]{$I^B$} (utility.north);
            \draw[dashed,->] ([xshift=0.5cm]utility.east)  -- node[midway,above]{$x$}(utility.east) ;
            \draw[->] (utility.west) -- node[midway,above]{$U$} (explorer.east);
            \draw[<->] (explorer.west) -- node[midway,above]{$B_r^x$} node[midway,below]{$S$} (navmanager.east);
            \draw[<->] (navmanager.west) -- node[midway,above]{$\eta$} node[midway,below]{$u$} (robot.east);
            \draw[dashed,->] ([yshift=0.5cm]navmanager.north)  -- node[midway,left]{$\mathcal{L},\mathcal{M}$} node[midway,right]{$x$}(navmanager.north) ;
            
		\end{tikzpicture}\quad
	}
	\caption[Proposed exploration framework]{Proposed exploration framework with our contribution in the red cloud. The Omnimapper SLAM system provides a local map $\mathcal{L}$, a global map $\mathcal{M}$ and robot localization $x$. Then, frontiers $f$ and perceived occupancy $\mathcal{W}$ are extracted from $\mathcal{M}$. The frontiers $f$ are clustered ($F$) and Behavioral entropy $\supscr{H}{B}$ is calculated from $\mathcal{W}$. For each cluster in $F$, a behavioral information gain $I^B$ is obtained from $\supscr{H}{B}$ that corresponds to the measurement model. Then, utilities $U$ are calculated from the information $I^B$ and pose estimate $x$. The explorer then picks a suitable goal region $B^x_r$ and sends it to the navigation manager, which in turn sends global and local plans $\eta$ to the controller.
 }\label{fig:framework}
\end{figure}
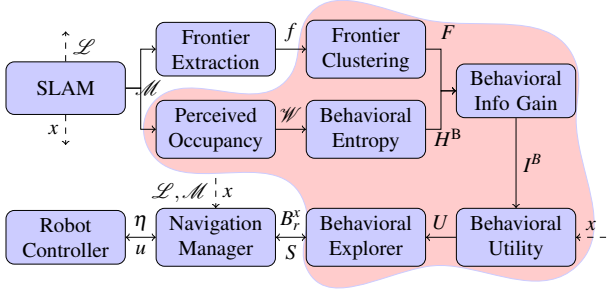

The framework is visualized in Fig.~\ref{fig:framework} and will be explained in detail in Sec.~\ref{sec:behavior_exploration}.

\section{Behavioral entropy and its analysis}
\label{sec:entropy}
We derive a new measure of uncertainty, $\supscr{H}{B} : \mathcal{P} \rightarrow \realnonnegative$, which we term Behavioral entropy. 
This function aims to quantify the subjectively perceived randomness present in a probability distribution.

\begin{figure}
    \begin{subfigure}[b]{0.48\linewidth}
        \includegraphics[width=\linewidth,height=2.2cm]{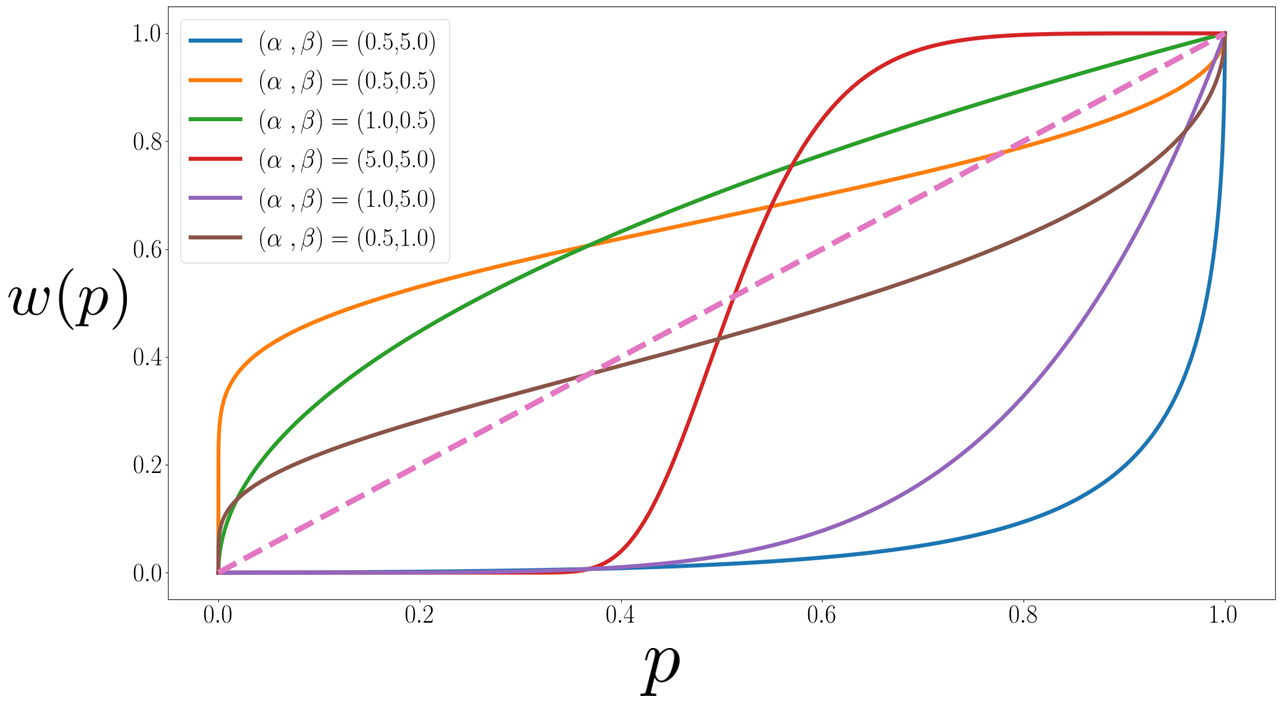}
        \caption{Prelec's weighting function}
        \label{fig:prelec_perception}
    \end{subfigure}
    \centering
    ~
    \begin{subfigure}[b]{0.48\linewidth}
        \centering
        \includegraphics[width=\linewidth]{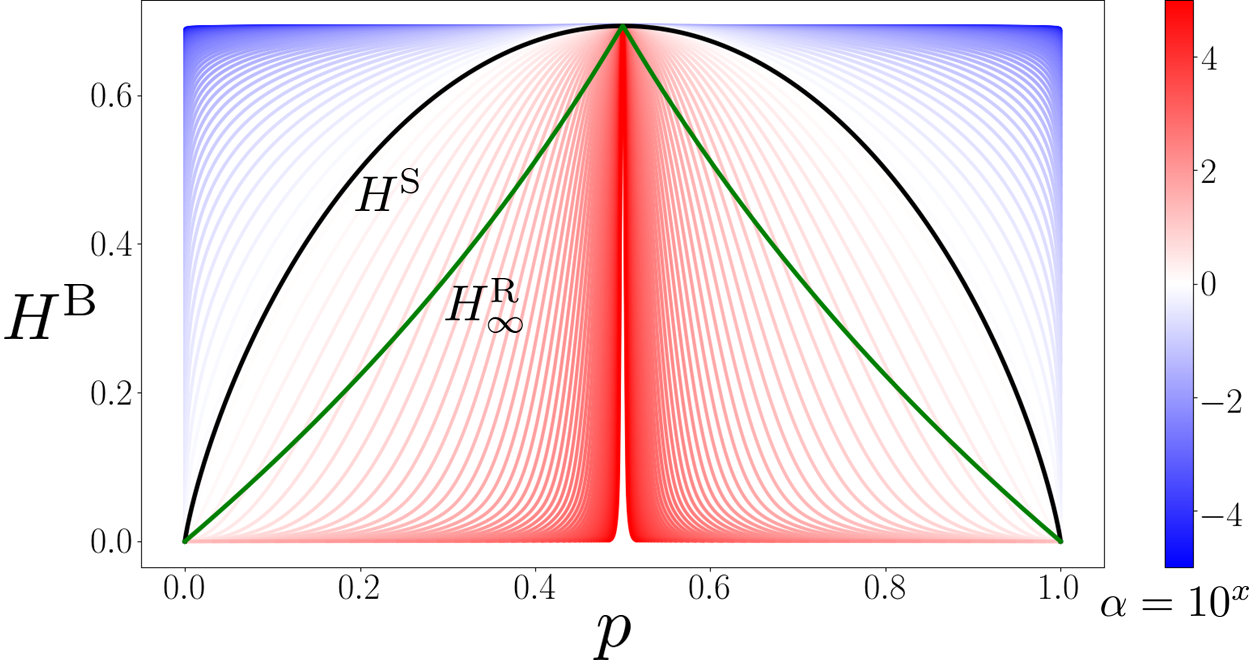}
        \caption{Entropy for Bernoulli trials.}
    \label{fig:entropy}
    \end{subfigure}
    \caption{(Left) Prelec's probability weighting function with a few parameter choices. Y-axis $w(p)$ indicates the perceived uncertainty associated with probability $p$ (X-axis). Dotted line indicates the unity curve $w(p)=p$. (Right) Behavioral entropy variation with $\alpha$ in log scale for a Bernoulli trial. Black curve indicates Shannon's entropy. }
    \label{fig:prob_ent}
\vspace{-0.5cm}
\end{figure}

Consider a probability vector $p = (p_1,\dots,p_M) \in \mathcal{P}_M$, and let $w : [0,1] \rightarrow [0,1]$ be \textit{a probability weighting function}, which maps $p_i \mapsto w(p_i) \equiv w_i$ and transforms the $\supscr{i}{th}$ outcome $p_i$ into a perceived probability $w_i$. Among the several probability weighting functions from the literature~\cite{AT-DK:92,SD:16}, we employ the popular Prelec's model~\cite{DP:98}: 
\begin{equation}
\label{eqn:prelec}
w(p)=e^{-\beta(-\log p)^{\alpha}}, \quad 
\alpha>0, \; \beta>0, \; w(0)=0, 
\end{equation}
see some examples in Fig.~\ref{fig:prelec_perception}. 
The parameter $\alpha$ controls the convexity or concavity of $w$: taking $\alpha \approx 0$ results in a highly concave $w$ that over-weights probabilities (i.e $w(p)\gg p$), while taking $\alpha \rightarrow \infty$ makes of $w$ a highly convex function that under-weights probabilities (i.e $w(p)\ll p$). The parameter $\beta$ controls the unique fixed point $p^*$ of $w$, i.e. the intersection with $w(p)=p$, the dotted line in Fig.~\ref{fig:prelec_perception}. 
A $\beta \approx 0$ indicates fixed point $p^* \approx 0$, while a $\beta \rightarrow \infty$ brings $p^* \approx 1$. By varying $\alpha,\beta$, we can model different behaviors. 
With low $\alpha$ and $\beta$ values results into an ``uncertainty averse'' behavior (green and orange curves in Fig.~\ref{fig:prelec_perception}), with $w(p)>p$ which implies more certainty in unlikely outcomes. With high $\beta$ values  we get ``uncertainty insensitive'' behavior when $w(p)<p$ (blue and brown curves in Fig.~\ref{fig:prelec_perception}), implying that the {\color{black}decision maker} only considers more certain outcomes. For more properties of these weighting functions, see~\cite{SD:16}. 
\subsection{Behavioral entropy and its analysis}
Our proposed ``Behavioral Entropy'' $\supscr{H}{B}$ combines the notion of 
 BGS entropy with Prelec's weights $w$ from \eqref{eqn:prelec}. 

\begin{equation}
\label{eqn:entropies_perceived}
    \supscr{H}{B}(p_1,\dots,p_M) = -\sum_{i=1}^M w(p_i) \log(w(p_i)). 
\end{equation}

Thus, incorporating the behavioral modelling capacity of Prelec's weighting function into an entropic function.
To analyze the new operator, we first list some  properties of Prelec's probability weighting functions $w$.

\begin{proposition}[Perceived probability~\cite{SD:16}, Prop.~$2.11$]
\label{prop:prelec_properties}
The class of weighting functions $w$ in~\eqref{eqn:prelec} satisfies: 
\begin{enumerate}
    \item $w$ is continuous, strictly increasing, and $w(1)=1$. 
    \item In addition to $0$ and $1$, $w$ only has another fixed point at $p^*=~e^{-(\frac{1}{\beta})^{\frac{1}{\alpha-1}}} \in (0,1)$. 
    $\qed$
\end{enumerate}
\end{proposition}

We use Proposition~\ref{prop:prelec_properties} to control fixed points. Specifically, to obtain $p^*=1/M$ for any integer $M \geq 2$, we substitute $p^*=\frac{1}{M}$ in 2) above and simplify terms to get conditions on $\alpha, \beta$ that guarantee $p^*=1/M$:
\begin{equation}
    \label{eqn:fixed_point_relation}
    \beta=e^{(1-\alpha)\log(\log(M))}. 
\end{equation}

Now we recall the following well-known result. 
\begin{lemma}[Maximality of $\supscr{H}{S}$]
\label{lem:shannon_maximality}
The BGS entropy function $\supscr{H}{S}$ attains its unique global maximum at $p \triangleq (\frac{1}{M},\dots, \frac{1}{M})$ (uniform distribution) over $\mathcal{P}_M$. $\qed$
\end{lemma}

Next, we show that $\supscr{H}{B}$ is admissible as a generalized entropy according to Definition~\ref{def:generalized_entropy} under \eqref{eqn:fixed_point_relation}.


\begin{theorem}
\label{thm:maximality}
Fix $M \in \natural$. If $\alpha,\beta$ satisfy~\eqref{eqn:fixed_point_relation}, then the function $\supscr{H}{B}$ is admissible as a generalized entropy. 
\end{theorem}

\begin{proof}
We need to show axioms A1, A2 and A3 are satisfied by $\supscr{H}{B}$.
Axioms A1 and A3 are trivially satisfied from the properties of $w$ in Proposition~\ref{prop:prelec_properties} and the fact that $\supscr{H}{S}$ satisfies both A1 and A3.
Regarding, A2, we observe that $\supscr{H}{B}(\mathbf{p}) = \supscr{H}{S}(\mathbf{w})$, where $\mathbf{p} = (p_1,\dots,p_M) \in \mathcal{P}_M$ and  $\mathbf{w} = (w_1,\dots,w_M)$ is the corresponding Prelec weights~\eqref{eqn:prelec}. From Lemma~\ref{lem:shannon_maximality}, {\color{black}$\supscr{H}{S}$} is maximized only with $\mathbf{w}$ with $w_i=\frac{1}{M}$, for all $i$. 
Then for any $\alpha,\beta$ satisfying~\eqref{eqn:fixed_point_relation}, the non-trivial fixed point of $w$ is necessarily $1/M$ and is unique. Hence $w_i=\frac{1}{M}$, for all $i$ only when $p_i=\frac{1}{M}$, for all $i$. Thus, $\supscr{H}{B}$ is maximized only when $\mathbf{p}$ is the uniform distribution.
\end{proof}

\begin{remark}
The Prelec's function~\eqref{eqn:prelec} with conditioning~\eqref{eqn:fixed_point_relation} allows us to control the fixed point of the function which is critical for admissibility guarantees, as well as the shape of the perceived probability curve, ensuring different behavioral perceptions. 
Whereas, other popular probability weighting functions~\cite{SD:16} like Karmakar's (fixed point is fixed at $0.5$) or Tversky and Kahneman's (shape and fixed point can't be controlled independently) cannot be used to generate meaningful and admissable generalized entropies.    \oprocend
\end{remark}

Next, we will compare the properties and performance of Behavioral entropy with other entropic functions. 
\subsection{Entropy comparison}
\label{sec:entropy_comparison}
First, we introduce the metrics that we use to make comparisons of different entropies. 

\begin{definition}[Sensitivity]
    \label{def:ent_sensitivity}
    Fix any generalized entropy function $H$ and assume that $H$ is normalized so that it is bounded above by $\log(M)$ (the maximizer of $H^S$). Then, $S(H) \in \realnonnegative$, the entropic sensitivity of $H$, is defined as
    \begin{equation}
    \label{eqn:entropic_sensitivity}
        S(H) = \int_{\mathcal{P}_M} H(\mathbf{p})d\text{Vol},
    \end{equation}
\end{definition}
where Vol is the standard Lebesgue measure over  $\mathcal{P}_M$. 
The value $S(H)$ provides a measure of the average sensitivity of $H$ to uncertainty over all distributions $\mathbf{p} \in \mathcal{P}$. 
If $S(H_1)>S(H_2)$, then the overall randomness perceived by $H_1$ is greater than that of $H_2$, and $H_1$ is more sensitive towards uncertainty.


\begin{definition}[Perceptiveness]
    \label{def:ent_perceptive}
    Let $H_\theta$ be any generalized entropy function as in Definition~\ref{def:ent_sensitivity} parametrized by $\theta \in \Theta \subseteq \real^A$. Let $H_\theta$ be continuous in $\theta$, then $P(H) \in \realnonnegative$, the perceptiveness of $H$, 
    is defined as:
    \begin{align}
    \label{eqn:entropic_perceptive}
        P(H) 
        =& \max_{\theta \in \Theta}S(H_\theta) - \min_{\theta \in \Theta}S(H_\theta).
    \end{align}
\end{definition}
The above notion of perceptiveness captures the range of sensitivity towards uncertainty for a class of entropies $H_\theta$, by considering all possible $\theta \in \Theta$. 

Now, we evaluate the perceptiveness of the family of $\supscr{H}{B}$. We begin by identifying a useful bound.
\begin{lemma}
\label{lem:ent_perceptive_limit}
Consider a generalized entropy $H_\theta$, which is bounded above by some $b \in \realpositive$, for all $\theta \in \Theta$. Then, $S(H_\theta)$ lies in the interval $(0,\frac{b}{M-1!})$, where $M$ is the number of outcomes. Subsequently, $0\le P(H_\theta) \le \frac{b}{M-1!}$. 
\end{lemma}
\begin{proof}
The volume of the probability simplex $\mathcal{P}_M\subseteq \real^M$ consisting of $M$ outcomes is $\frac{1}{(M-1)!}$~\cite{PS:66-simplex}. 
The limits of $S(H_\theta)$ can be computed by considering that $0 \le \inf_{\theta \in \Theta} \int_{\mathcal{P}_M} H_\theta(\mathbf{p})d\text{Vol} \leq S(H_\theta) \leq \sup_{\theta \in \Theta} \int_\mathcal{P} H_\theta(\mathbf{p})d \text{Vol}$. 
\end{proof}

With this let us look at the perceptiveness of $\supscr{H}{B}$.

\begin{theorem}
\label{thm:per_ent_perceptiveness}
Consider $\supscr{H}{B}_{\alpha,\beta} \equiv \supscr{H}{B}_{\alpha}$ conditioned according to~\eqref{eqn:fixed_point_relation}. The sensitivity $S(\supscr{H}{B}_\alpha)$ lies in the interval $(0,\frac{\log(M)}{M-1!})$ its perceptiveness $P(\supscr{H}{B})$ reaches the maximum possible  $\frac{\log(M)}{M-1!}$.
\end{theorem}
\begin{proof}
We will first evaluate the limits of $\supscr{H}{B}$ in $\alpha$ and then apply it to sensitivity measurement $s$.
We note that $\supscr{H}{B}$ is continuous in $\alpha$ and has a unique maximum at the uniform distribution $\subscr{\mathbf{p}}{uni}$ when conditioned according to~\eqref{eqn:fixed_point_relation}. The limits of $S(\supscr{H}{B})$ can be calculated as follows:
\begin{equation}
    s_0=\lim_{\alpha \rightarrow 0}S(\supscr{H}{B}_\alpha)=\lim_{\alpha \rightarrow 0} \int_{\mathcal{P}_M} \supscr{H}{B}_\alpha(\mathbf{p})d\text{Vol}.
\end{equation}
Since $\supscr{H}{B}$ is continuous and bounded above by $\log(M)$ for every $P \in \mathcal{P}$, and the volume of $\mathcal{P}$ region is finite, 
we can take the limit inside the integral and obtain:
\begin{align*}
    s_0=&\int_{\mathcal{P}_M} \lim_{\alpha \rightarrow 0}  \supscr{H}{B}_\alpha(P)dP \\
    =&\int_{\mathcal{P}_M} \sum_{i=1}^M - \lim_{\alpha \rightarrow 0}(e^{-\beta_\alpha(-\log(p_i))^\alpha} \beta_\alpha(-\log(p_i))^\alpha))\\
    =& \int_{\mathcal{P}_M} M \times \frac{1}{M} \times \log(M) = \frac{\log(M)}{M-1!}.
\end{align*}

Next for the lower bound, 

\begin{align*}
    s_\infty=&\int_{\mathcal{P}_M} \lim_{\alpha \rightarrow \infty}  \supscr{H}{B}_\alpha(P)dP \\
    =&\int_{\mathcal{P}_M} \sum_{i=1}^M - \lim_{\alpha \rightarrow \infty}(e^{-\beta_\alpha(-\log(p_i))^\alpha} \beta_\alpha(-\log(p_i))^\alpha))\\
    =& \int_{\mathcal{P}_M} M \times 0 \times \log(0) = 0,
\end{align*}
and the result follows.
\end{proof}
Now we compare $\supscr{H}{B}$ with Renyi and Shannon entropies on the basis of perceptiveness.
\begin{theorem}
\label{thm:perceptiveness_comparison}
Consider the family of $\subscr{H}{B} \equiv (\supscr{H}{B}_\alpha)$ conditioned according to~\eqref{eqn:fixed_point_relation}. Then, $P(\supscr{H}{B}) > P(\supscr{H}{R}) > P(\supscr{H}{S})$.
\end{theorem}
\begin{proof}
From ~\eqref{eqn:entropies}, the value $P(\supscr{H}{R})$ can be obtained from
appropriate limits of $S(\supscr{H}{B})$.
The upper limit of Renyi sensitivity is reached when $\gamma \rightarrow 0$ and it is easy to see that coincides with that of Behavioral entropy:
\begin{equation}
    s_0=\lim_{\gamma \rightarrow 0} \int_{\mathcal{P}_M} \supscr{H}{R}_\gamma(\mathbf{p})d\text{Vol}= \frac{\log(M)}{M-1!}.
\end{equation}
For the lower bound, since Renyi entropy is decreasing in $\gamma \in (0,1)\cup(1,\infty)$, the lowest sensitivity is reached when 
\begin{align*}
    s_\infty=&\int_{\mathcal{P}_M} \lim_{\gamma \rightarrow \infty}  \supscr{H}{R}_\gamma(\mathbf{p})d\text{Vol} 
    =\int_{\mathcal{P}_M} \hspace*{-0.4cm}- \log (\max\{p_1,\dots,p_M\}) d\text{Vol}.
\end{align*}
Since $0 < \max_{\mathbf{p} \in \mathcal{P}_M}\{p_1,\dots,p_M\} \le 1$, as $\mathbf{p}$ is a probability distribution, then $- \log (\max_{\mathbf{p} \in \mathcal{P}_M}\{p_1,\dots,p_M\})\ge 0 $ for all $\mathbf{p}$. Hence, the lower limit $s_\infty(\supscr{H}{R}) \ge 0$.
Thus, we get $P(\supscr{H}{B}) > P(\supscr{H}{R})$
 from Theorem~\ref{thm:per_ent_perceptiveness}. 
 Next it is trivial to show that $P(\supscr{H}{R}) > P(\supscr{H}{S})$ as the perceptiveness of $\supscr{H}{S}$  is $0$. 
\end{proof}

The concept of perceptiveness and its comparison with different entropies is visualized in Fig.~\ref{fig:entropy} using Bernoulli trials. Behavioral entropy $\supscr{H}{B}$ is able to capture the entire behavior spectrum from ``Uncertainty Averse'' (blue region) to ``Uncertainty ignorant''  (red region). 
Shannon entropy $\supscr{H}{S}$ is a single curve (black). Renyi entropy $\supscr{H}{R}$ can capture ``Uncertainty Averse'' (blue region) behavior however, it can't be more ``uncertainty ignorant'' than $\supscr{H}{R}_\infty$, thus being significantly less perceptive than $\supscr{H}{B}$. Next, we connect the Behavioral entropy notion to a robotic exploration setting.

\section{Behavioral Exploration}
\label{sec:behavior_exploration}
Here, we provide details on the frontier selection and clustering process for robotic exploration, and explain the construction of the utility function used to choose the best frontier, which defines our framework in Fig.~\ref{fig:framework}.

\paragraph*{Frontier Extraction and Clustering}
The known areas $\supscr{\mathcal{M}}{K}$, unknown areas $\supscr{\mathcal{M}}{UK}$, and uncertain areas $\supscr{\mathcal{M}}{U}$ of the occupancy map $\mathcal{M}$ are extracted from the occupancy values as described in Sec.~\ref{sec:prelims}. 
Then, 
the free space $\supscr{\mathcal{M}}{F}$, resp.~the occupied space $\supscr{\mathcal{M}}{O}$, is obtained by collecting cells whose occupancy probability is less than $\supscr{\tau}{FS}$, resp.~greater than $\supscr{\tau}{OB}$. 
Gradient vectors are computed over $\mathcal{M}$
and the non-zero cells are stored in $\mathcal{M}'$ defining all possible frontier candidates. 
Obstacle neighbor cells $\supscr{\mathcal{M}}{ON}$ are given by  the convolution of $\supscr{\mathcal{M}}{O}$ with a suitable kernel $\supscr{K}{N}$. 
Frontier cells $\mathcal{F}$ are the free cells 
that are part of the gradient 
but not neighbors of obstacles. 
As this can create a large number of frontier cells, these are clustered $\supscr{\mathcal{F}}{C}$ according to the cluster size
$\supscr{\tau}{CL} \in \integernonnegative$, which indicates the size of square grid that clusters frontiers into a single frontier. 
Greater $\supscr{\tau}{CL}$ provides lesser number of clusters while $\supscr{\tau}{CL}\triangleq 1$ results in no clustering. 
A representative frontier is picked at random in each cluster to give the final list of chosen frontiers $\supscr{\mathcal{F}}{L}$, whose size is equal to the number of clusters. This procedure is summarized in Algorithm~\ref{alg:cluster}.

\begin{algorithm}
\SetAlgoLined
Input: $\mathcal{M},\supscr{\tau}{OB},\supscr{\tau}{FS},\supscr{\tau}{CL},\supscr{K}{N}$ ;\\
Output : $\mathcal{F}^L$; \\
$\mathcal{M}^K,\ \mathcal{M}^{UK},\ \mathcal{M}^U \leftarrow \mathcal{M} $ \tcp*{\scriptsize Partition map}
$\mathcal{M}^F \leftarrow (\mathcal{M}^K \cup \mathcal{M}^U)< \supscr{\tau}{FS}$ \tcp*{\scriptsize Get free space}
$\mathcal{M}^O \leftarrow (\mathcal{M}^K \cup \mathcal{M}^U)> \supscr{\tau}{OB}$ \tcp*{\scriptsize Get occupied space}
$\mathcal{M}' \leftarrow \mathcal{M} $ \tcp*{\scriptsize map areas with non-zero gradients}
$\mathcal{M}^{ON} \leftarrow \mathcal{M}^O \circ \supscr{K}{N} $ \tcp*{\scriptsize obstacle neighbor cells}
$\mathcal{F} \leftarrow (\mathcal{M}^F \cap \mathcal{M}') - \mathcal{M}^{ON} $\tcp*{\scriptsize free gradient cells} 
$\mathcal{F}^C \leftarrow \text{Cluster}(\mathcal{F}, \supscr{\tau}{CL}) $\tcp*{\scriptsize cluster frontiers}
$\mathcal{F}^L \leftarrow \text{Random Pick}(\mathcal{F}^C)$
\caption{Frontier extraction and clustering}\label{alg:cluster}
\end{algorithm}

\paragraph*{Information Gain and Utility function}
A sensor footprint $\mathcal{A} \subset \mathcal{M}$ is obtained around each $f\in \mathcal{F}$, which is a collection of occupancy cells according to the sensor model.
We use the LiDAR beam model~\cite{ST-WB-DF:05} with range and number of beams to determine the cells {\color{black}that} get observed. 
We then update the occupancy for each cell in $\mathcal{A}$ based on the number of beams that hit a cell. 
Next, a perceived occupancy is calculated using~\eqref{eqn:utility_frontier}. Here, the behavioral information gain $\supscr{I}{B}$ is calculated as the reduction in entropy
from observing the area $\mathcal{A}$ around $f$.
We then use the following utility function to determine the value $u \in \realnonnegative$ of going to $f'$ from the current pose $x$ and map~$\mathcal{M}$:

\begin{equation}
    \supscr{u}{B}(f',x;\mathcal{M})= \frac{\supscr{I}{B}(f';\mathcal{M})}{|\eta(x,f')|}, \ \supscr{I}{B}(f';\mathcal{M})= \sum_{x' \in \mathcal{A}}\supscr{H}{B}(p((x'))
    \label{eqn:utility_frontier}
\end{equation}
Here, $\supscr{I}{B}(f';\mathcal{M})$ is the information gained or entropy reduced by observing area $\mathcal{A}$ around frontier $f'$ based on the current occupancy map $\mathcal{M}$. $p((x'))$ indicates the Bernoulli occupancy distribution at $x'$. 
$|\eta(x,f')|$ is the path length from position $x$ to frontier $f'$.
The frontier with the highest utility is chosen to visit; see Algorithm~\ref{alg:info_gain}. 
\begin{equation}
    f^*=\argmax_{f' \in \mathcal{F}^L} \supscr{u}{B}(f',x;\mathcal{M}).
    \label{eqn:frontier_selection} 
\end{equation}

\begin{lemma}
    Consider the entropies $\supscr{H}{S}$, $\supscr{H}{R}$ and $\supscr{H}{B}$ conditioned according to~\eqref{eqn:fixed_point_relation}, the corresponding utilities $\supscr{u}{S}$, $\supscr{u}{R}$, and $\supscr{u}{B}$. For any given map $\mathcal{M}$, pose $x$ and frontier location $f'$, the range $\mathcal{R}(\supscr{u}{x}) \in \realpositive$ of possible utilities $\supscr{u}{x}$ follows the order $\mathcal{R}(\supscr{u}{B}) > \mathcal{R}(\supscr{u}{R}) > \mathcal{R}(\supscr{u}{S})$   
\end{lemma}
\begin{proof}
The proof trivially follows from Theorem~\ref{thm:perceptiveness_comparison}
\end{proof}

\begin{remark}
    As utilities calculated from Behavioral entropy $\supscr{H}{B}$ have the highest range, chosen frontier $f^*$ from \eqref{eqn:frontier_selection} {\color{black} with different parameter choices} are also the most diverse. For instance, frontiers with high quality information will be valued higher with high $\alpha$ irrespective of path length, forcing aggressive exploration farther away. On the other end, a low $\alpha$ will treat any uncertainty almost equally, thus focusing exploration in nearby areas. 
\end{remark}

The proof of concept (POC) simulations in Sec.~\ref{sec:poc_sim} provides more insight.
We describe the exploration pipeline (Fig.~\ref{fig:framework}).
\begin{algorithm}
\SetAlgoLined
Input: $\mathcal{M},\mathcal{F}^L,\alpha, x $ ;\\
Output : $f^*$; \\
\For{$f \in \mathcal{F}^L$}{
    $\mathcal{A} \leftarrow f, \mathcal{M}$ \tcp*{\scriptsize predicted sensor footprint}
    $\mathcal{W} \leftarrow \mathcal{A}, \alpha$ \tcp*{\scriptsize perceived probabilities}
    $\supscr{H}{B} \leftarrow \mathcal{W}$ \tcp*{\scriptsize Behavioral entropy}
    $I \leftarrow \supscr{H}{B}$ \tcp*{\scriptsize information gain}
    $|\eta| \leftarrow \text{planner}(f,x)$ \tcp*{\scriptsize path length}
    $u \leftarrow \text{utility}(I,|\eta|)$ \tcp*{\scriptsize frontier utility}
} $f^* \leftarrow \text{argmax}U$ \tcp*{\scriptsize most informative location}
\caption{Behavioral frontier selection}\label{alg:info_gain}
\end{algorithm}

\paragraph*{Exploration pipeline}
Algorithm~\ref{alg:explorer} describes the behavioral exploration framework in Fig~\ref{fig:framework}. 
The robot starts with an initial pose $x$. The observations at $x$, $Y$, are used to construct the occupancy map $\mathcal{M}$ using SLAM. 
Then, the frontier list $\mathcal{F}^L$ is extracted by Algorithm~\ref{alg:cluster} and the best frontier $f^*$ is selected according to Algorithm~\ref{alg:info_gain}. 
Then, a circular goal region $\mathbf{B}_r^{x^*}$ is computed, where $x^*$ is the transformed pose of $f^*$ in world frame and $r$ is radius of the region. 
The navigation manager then provides appropriate global and local plans $\eta$ to the goal region $\mathbf{B}_r^{x^*}$, to generate a control policy $\Pi$ that ensures this is reached. 
SLAM, navigation manager, controller and frontier clustering are continually running in the background to ensure accurate and updated information. 
The frontier selection process updates the new frontier $f^*$ once the navigation manager either confirms that the goal is reached or aborts when no feasible plans are found. 
The algorithm runs until there are no more frontiers to explore or the perceived information gain is negligible ($\approx 0$) from remaining frontiers.

\begin{algorithm}
\SetAlgoLined
Input: $\mathcal{M}, x, K$ ;\\
Output : $\Pi$; \\
Initialise : $\mathcal{M}, x ,r, \mathcal{F}^L $;\\
Initialise : $ x ,r$;\\
\While{$\mathcal{F}^L \neq \phi \ \& \ \supscr{I}{B} \neq \mathbf{0}$}{
$\mathcal{M},x \leftarrow \text{SLAM}(Y,\Pi) $ 
$\mathcal{F}^L \leftarrow \text{Frontier Cluster}(\mathcal{M},K) $ 
$f^* \leftarrow \text{Frontier Selection}(\mathcal{M},\mathcal{F}^L,x,K) $ 
$x^* \leftarrow \text{Pose transform}(f^*,\mathcal{M})$ 
$r \leftarrow \text{Goal Radius}(\mathbf{B}^p_r)$ 
$\mathbf{B}^{x^*}_r \leftarrow \text{Goal region}(r,x^*,\mathcal{M})$ 
$\eta \leftarrow \text{Navigation Manager}(\mathbf{B}^{x^*}_r,\mathcal{M})$ 
$\Pi \leftarrow \text{Robot Controller}(\eta,x,\mathcal{M}) $ 
}
\caption{Behavioral explorer}\label{alg:explorer}

\end{algorithm}

\section{Proof of concept simulations}
\label{sec:poc_sim}
To evaluate our proposed approach, we consider first a proof of concept (POC) environment and then focus on real world scenarios in Sec.~\ref{sec:exp_results}.
Here, we assume all other aspects of exploration: SLAM, navigation and controls are working optimally to focus exclusively on exploration.

\begin{figure}
    \begin{subfigure}{0.48\linewidth}
        \centering
        \includegraphics[width=\linewidth]{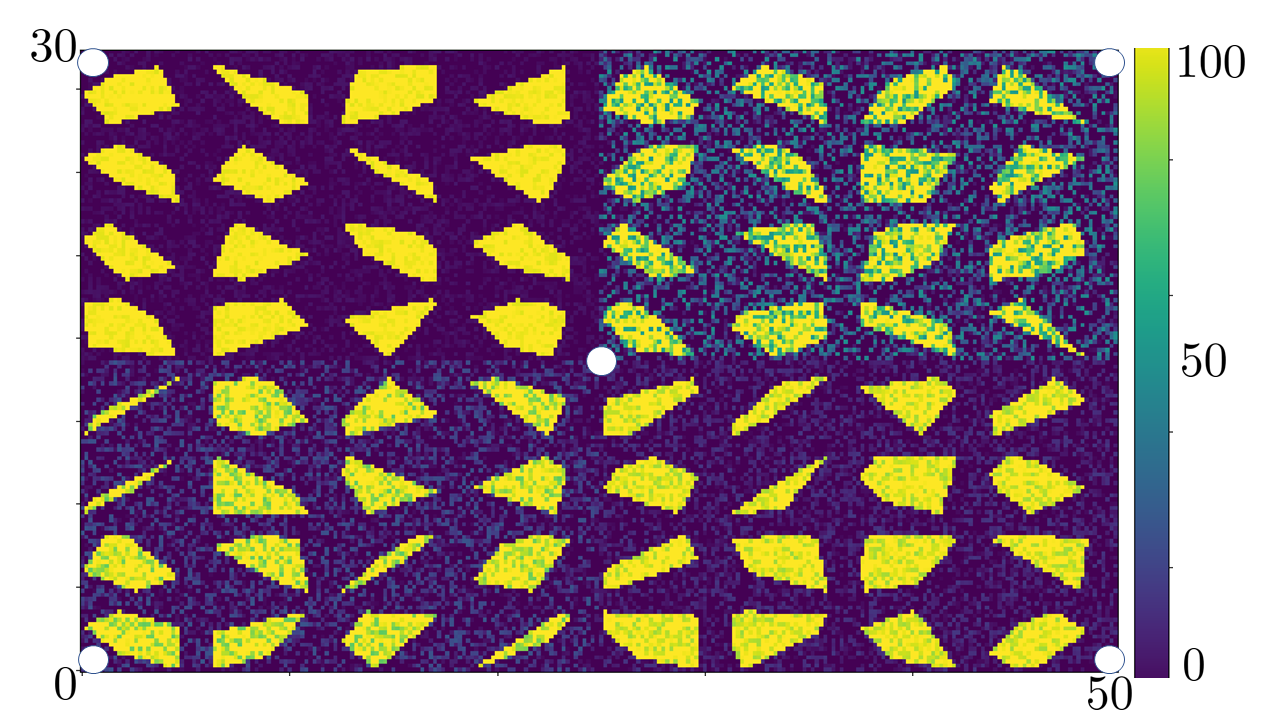}
        \caption{Environment used for POC simulation.}
        \label{fig:poc_env}
    \end{subfigure}
    ~
    \begin{subfigure}{0.48\linewidth}
        \centering
        \includegraphics[height=0.8\linewidth, angle=90]{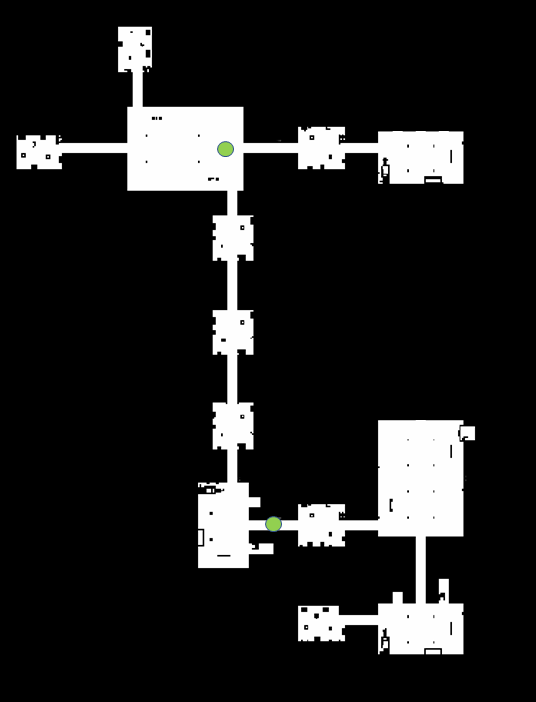}
        \caption{The ground truth map of the DARPA SubT section used.}
        \label{fig:ground_truth}
    \end{subfigure}
    \caption{Environments used for simulations}
    \label{fig:sim_environment}
    \vspace{-0.5cm}
\end{figure}

\paragraph*{Environment Setup}
\label{sec:poc_setup}
\label{sec:environment_setup}
Consider a 2D rectangular environment $\mathcal{X}=[0,30] \times [0,50] $ (Fig~\ref{fig:poc_env}) discretized by $0.1$ unit square grid cells ($\mathcal{D}=[0,300] \times [0,500]$) containing randomly shaped polygonal obstacles (yellow/bright).  
The ground truth map $\mathcal{M}^{*}$ consists of free-space ($0$ occupancy value) and obstacles ($100$ occupancy value). 
The initial occupancy map $\mathcal{M}$ is constructed by adding noise to each cell value by 1) sampling from uniform distribution over each quadrant, 2) adding those to ground truth values of free-space cells ($0$) and subtracted from obstacle cells ($100$), see Fig~\ref{fig:poc_env} where this is shown. 
Blue/dark areas ($0$) represent free space, bright/yellow areas ($100$) represent obstacles and other values indicate uncertain areas. 
The environment is divided into four quadrants which contain different levels of noise, sampled from a uniform distribution with different intervals from $[0,5]$ (top-left quadrant), to $[0,50]$ (top-right), and others with $[0,15]$ (bottom-right) and $[0,25]$ (bottom-left). 
The white circles in the map are the five initial conditions for the robot, $x^0_i, i \in \{1,2,3,4,5\}$. From these, given initial map in Fig~\ref{fig:poc_env}, the robot explores the entire region via Algorithm~\ref{alg:explorer} to recover $\mathcal{M}^*$. 

\paragraph*{POC exploration pipeline}
First, the agent updates a circular area $\mathcal{A} = \mathbf{B}^x_r$ in $\mathcal{M}$, with a radius $r$ and centered around its current pose $x$. 
Four different mapping radii $\{2,3,4,5\}$ to capture different sensor ranges and 
three levels of mapping noise $\sigma^m \in \{0,1,2\}$ are employed with the following meaning: \textit{a)} In perfect mapping ($\sigma^m \triangleq 0$) each cell in $\mathcal{A}$ is updated to the ground truth.
{\color{black}\textit{b)} In imperfect mapping ($\sigma^m \triangleq 1$ and $\sigma^m \triangleq 2$) the occupancy values in $\mathcal{A}$ are reduced/increased randomly\footnote{with a random number generated by a uniform distribution with range $[0,35]$ for $\sigma^m \triangleq 1$ and $[0,15]$ for $\sigma^m \triangleq 2$ for each cell in $\mathcal{A}$}, with higher $\sigma^m$ implying lesser reduction in uncertainty after observation.}  

Second, frontiers $\mathcal{F}$ are obtained as those cells with occupancy values $<2$ and  a non-negative gradient w.r.t. to the cells around them.
We do not cluster the frontiers here to make sure all possible frontiers are considered. 

Then, for each $f_i \in \mathcal{F}$, we calculate the information gain by considering  $\mathcal{A}=\mathbf{B}^{f_i}_r$.
That is, for each cell in $\mathcal{A}$, the information gain per cell is given by the Behavioral entropy in these cells and the total information gain $I$ is the sum of all of these values. 
Then, for each $f_i$, the utility is calculated as the information gain per Euclidean path length required to reach $f_i$ from $x$. 
The best frontier $f^*$ is that with  the maximum information gain.
Finally, the robot moves from $x$ to $f^*$ in a straight line while mapping (first step) all possible circular regions within its sensor range along the path. 

\begin{figure}
     \centering
      \begin{subfigure}[b]{0.23\textwidth}
         \centering
         \includegraphics[width=\textwidth]{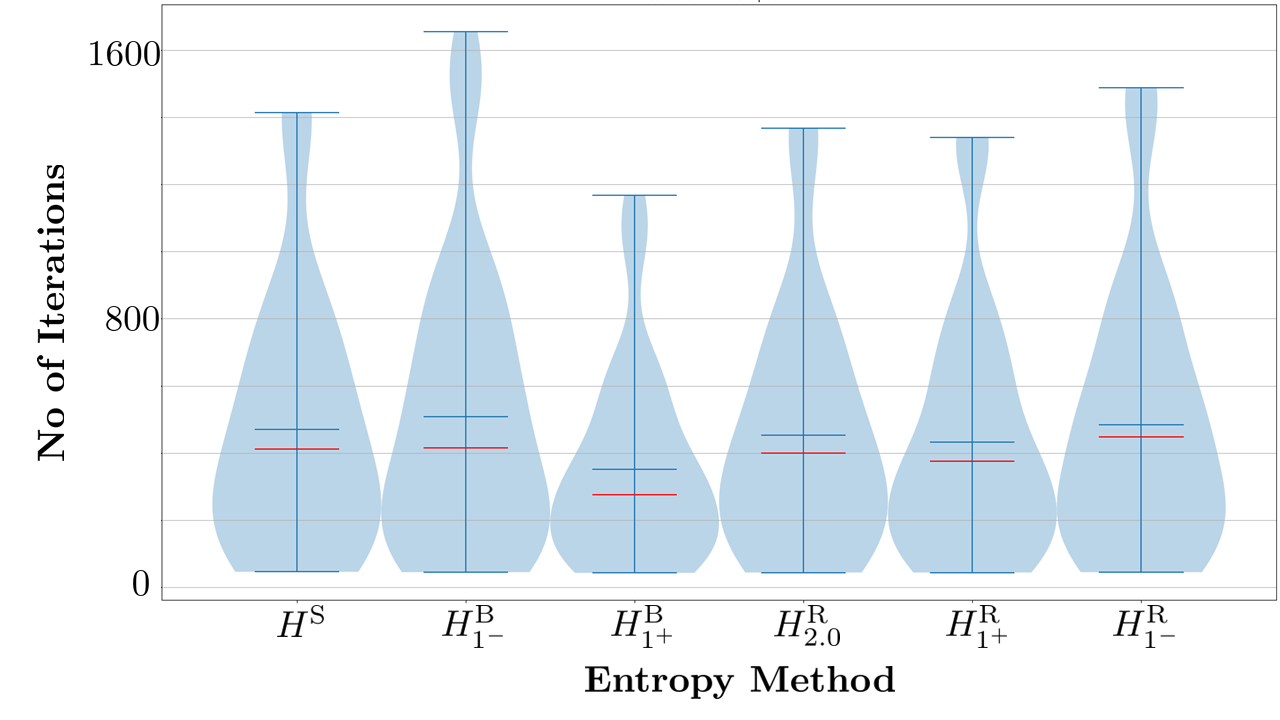}
         \caption{Entropy reduced all trials}
         \label{fig:results_poc_1}
     \end{subfigure}
     ~
     \begin{subfigure}[b]{0.23\textwidth}
         \centering
         \includegraphics[width=\textwidth]{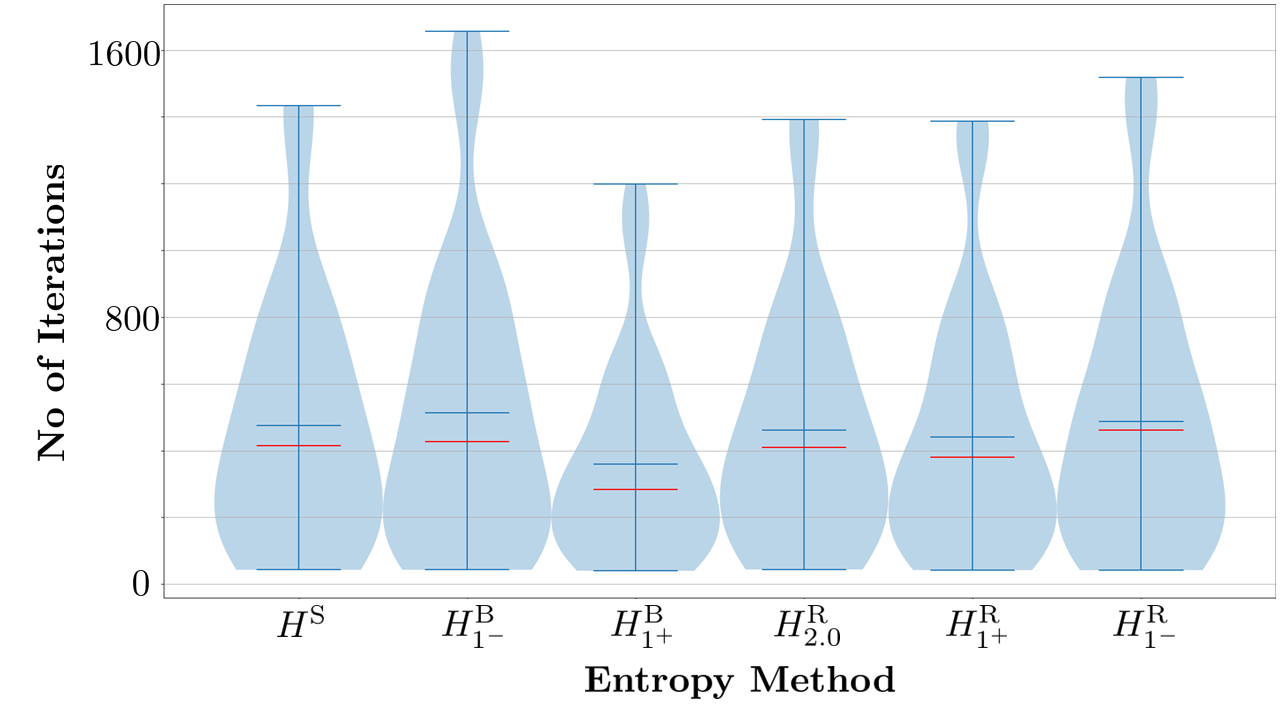}
         \caption{Area explored all trials}
         \label{fig:results_poc_2}
     \end{subfigure}
     
     \begin{subfigure}[b]{0.23\textwidth}
         \centering
         \includegraphics[width=\textwidth]{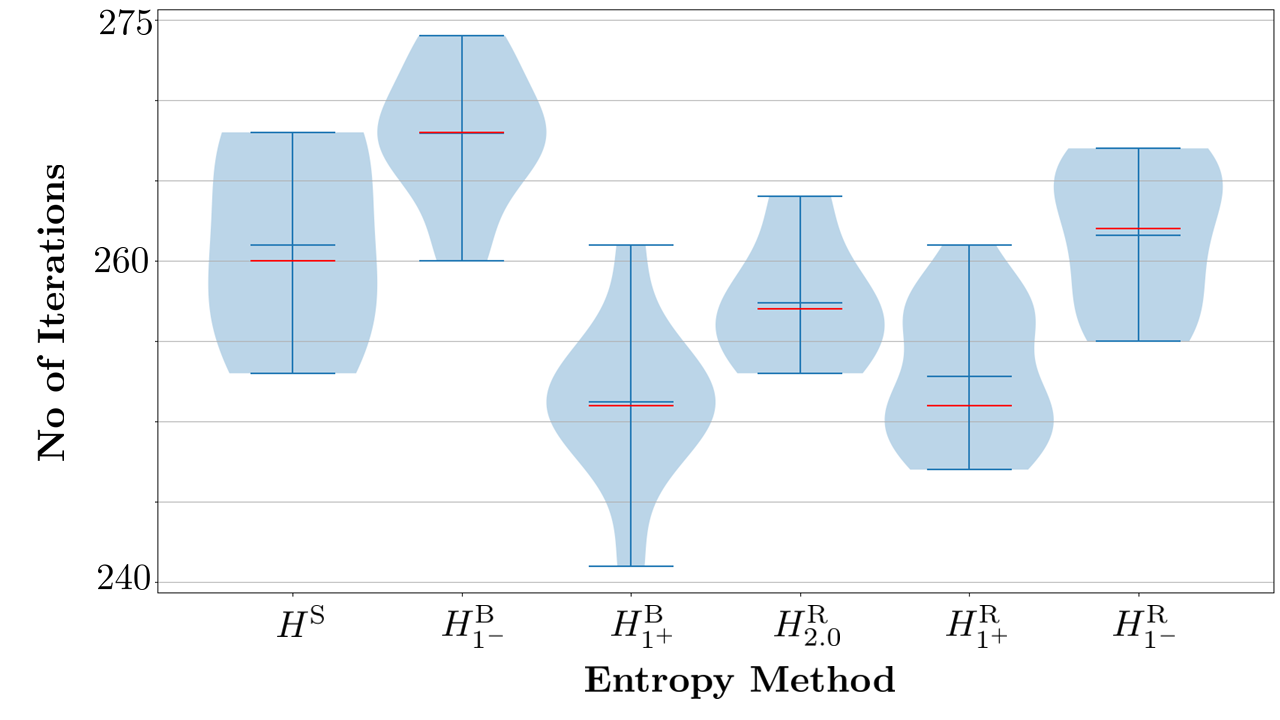}
         \caption{Entropy reduced ($r=2$,$\sigma^m=0$)}
         \label{fig:results_poc_3}
     \end{subfigure}
     ~
     \begin{subfigure}[b]{0.23\textwidth}
         \centering
         \includegraphics[width=\textwidth]{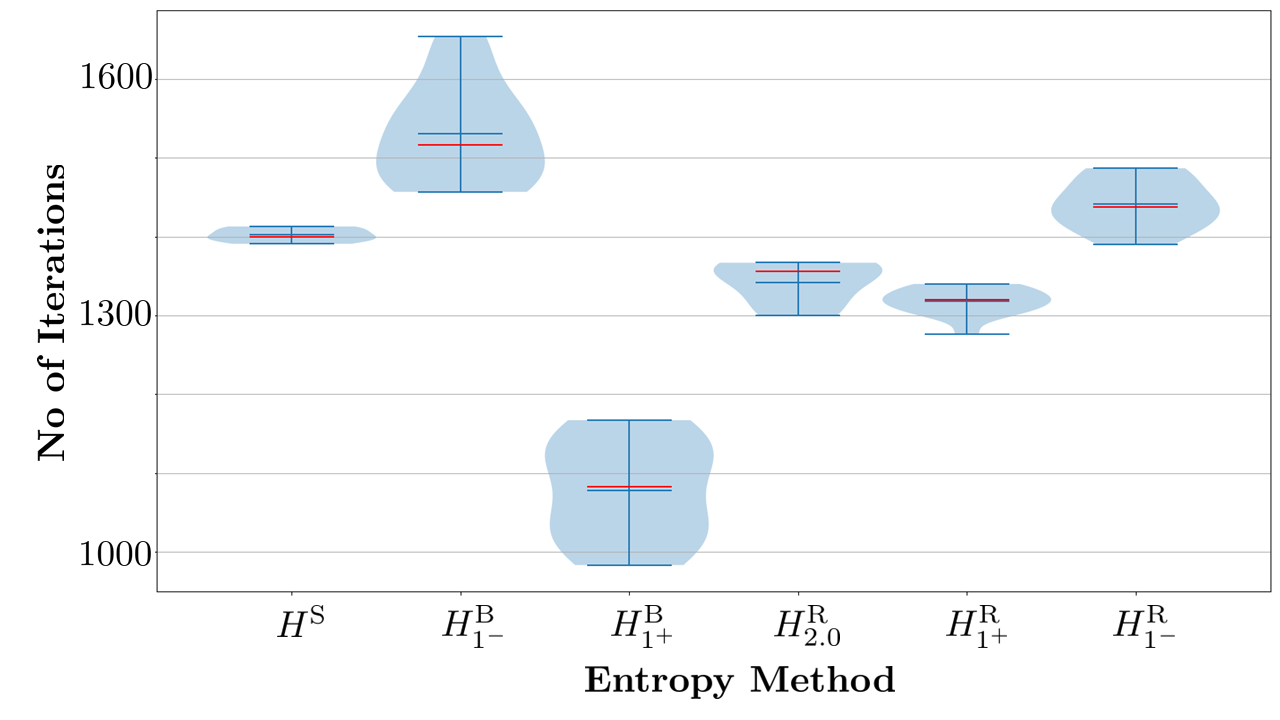}
         \caption{Entropy reduced ($r=2$,$\sigma^m=2$)}
         \label{fig:results_poc_4}
     \end{subfigure}
     
     \begin{subfigure}[b]{0.23\textwidth}
         \centering
         \includegraphics[width=\textwidth]{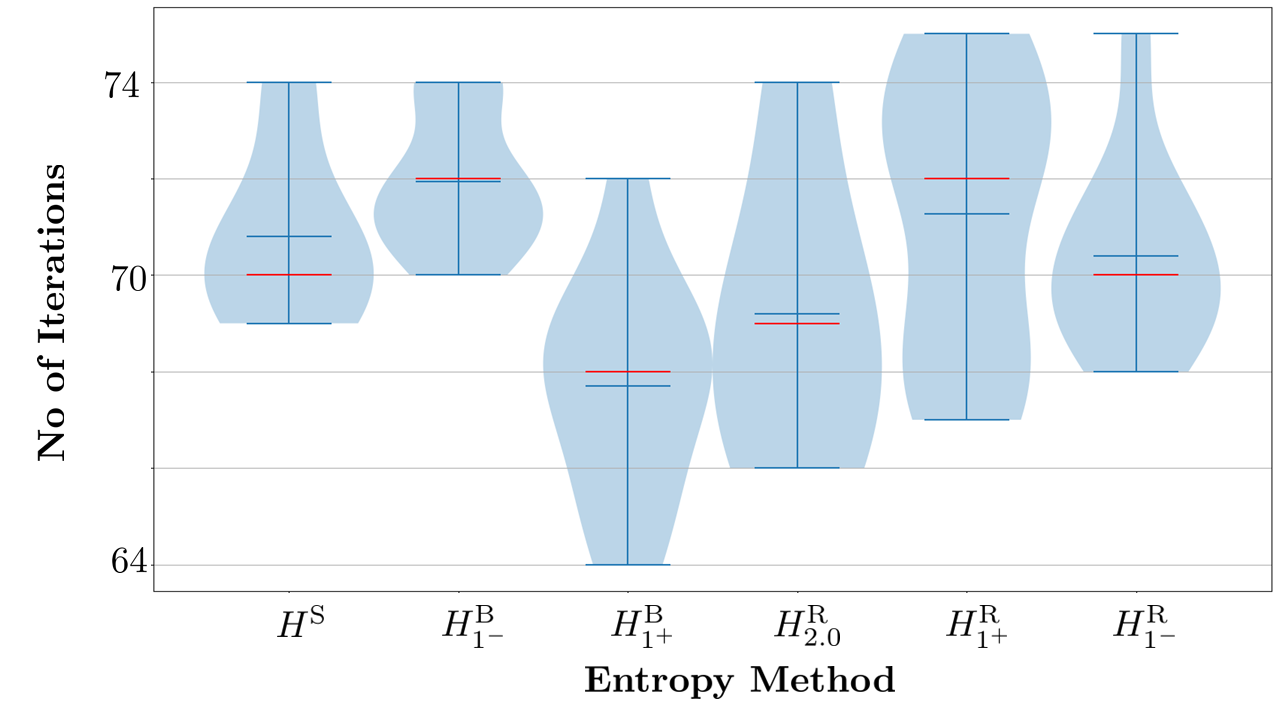}
         \caption{Entropy reduced ($r=4$,$\sigma^m=0$)}
         \label{fig:results_poc_5}
     \end{subfigure}
     ~
     \begin{subfigure}[b]{0.23\textwidth}
         \centering
         \includegraphics[width=\textwidth]{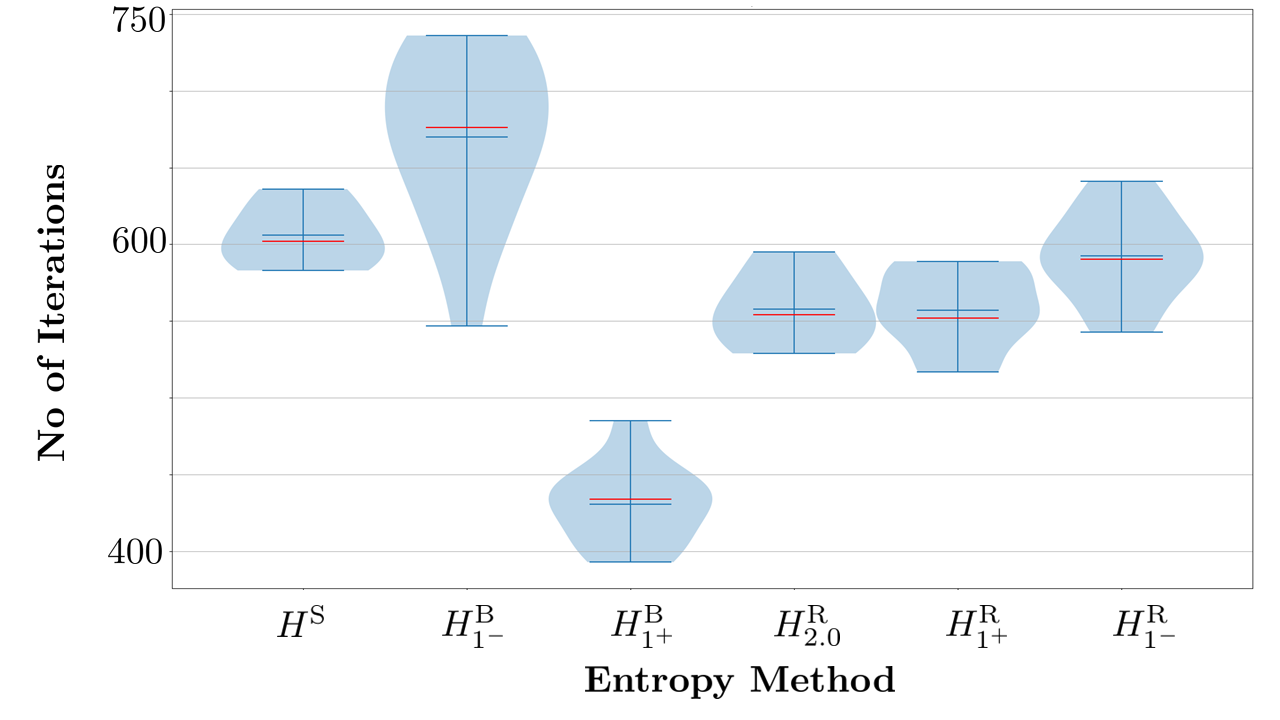}
         \caption{Entropy reduced ($r=4$,$\sigma^m=2$)}
         \label{fig:results_poc_6}
     \end{subfigure}
        \caption{ Violin plots indicating the number of iterations used to complete $99\%$ of exploration under different behavioral exploration strategies, sensing radius and noise conditions}
        \label{fig:results_poc}
        \vspace{-0.5cm}
\end{figure}

\paragraph*{Trials and Metrics}
In summary, we have the following control variables: a) sensor radius $r \in \{2,3,4,5\}$, b) mapping noise $\sigma^m \in \{0,1,2\}$, c) starting point $x^0_i, i \in \{1,2,3,4,5\}$, d) entropy method and parameter choice: $\alpha \in \{0.2,0.5,0.8,2.0,3.0,5.0\}$ for Behavioral entropy, $\gamma \in \{0.2,0.5,0.8,2.0,10.0,100.0,1000.0 \}$ for Renyi's entropy, and Shannon's entropy. Resulting in 60 trials for each entropy method. 
{\color{black}For the ease of comparison, we group the behavioral and Renyi parameters into ``Uncertainty averse'' $\subscr{\supscr{H}{B}}{1-}$,$\subscr{\supscr{H}{R}}{1-}$ corresponding to $\alpha,\gamma < 1$ and ``Uncertainty ignorant'' $\subscr{\supscr{H}{B}}{1+}$,$\subscr{\supscr{H}{R}}{1+}$ corresponding to $\alpha,\gamma > 1$ respectively.} 
We also consider Renyi quadratic entropy (RQE) with $\gamma=2$, as this is widely used in the literature~\cite{carrillo2018autonomous,charrow2015information}.  
We use the ground truth to measure the total uncertainty remaining in $\mathcal{M}$, which is given by Shannon's entropy for a fair comparison.  
After every iteration of Algorithm~\ref{alg:explorer}, we record the entropy remaining and the percentage of completion. 
We keep the same random number generator for sampling the mapping noise per trial to ensure accountability. 
\paragraph*{Results and Discussion}
\label{sec:poc_results}
\color{black}
The results in Fig~\ref{fig:results_poc} show the distribution of the number of iterations taken to explore $99\%$ of the map. The $Y$-axis indicates the distribution of number of iterations used with a particular group to complete $99\%$ of area/entropy. The largest and smallest number of iterations are indicated by blue horizontal lines limiting the violin shapes, while mean and median are given by horizontal blue and red lines inside the violin, respectively. 
The wider the violin, the more trials resulted in similar number of iterations, while the narrower the violin, the fewer the trials.

First, we report the performance for each entropy method group across all trials, see Fig~\ref{fig:results_poc}. We found that the results of area explored and entropy reduced
  have similar trends as seen in Fig~\ref{fig:results_poc_1} and Fig~\ref{fig:results_poc_2}. Henceforth, we will report and discuss both metrics as the same.
Fig~\ref{fig:results_poc_1} shows that the minimum time for full exploration is similar for all groups, resulting from trials with very good sensor capability ($r=4,5$) under perfect mapping $\sigma^m=0$. This is illustrated in Fig~\ref{fig:results_poc_5} for $r=4$, we see that across all behaviors, the number of iterations are very similar ranging from $64-75$. 
We also see a similar trend with $r=4$ and $\sigma^m=0$, see Fig.~\ref{fig:results_poc_3}. Thus with perfect mapping $\sigma^m=0$, all behavioral methods seem to provide similar results with marginally better performance from ``Uncertainty ignorant'' $\subscr{\supscr{H}{B}}{1+}$ behavior.
In general, from Fig.~\ref{fig:results_poc_1} and Fig.~\ref{fig:results_poc_2} mean and median iterations are significantly less for ``Uncertainty ignorant'' Behavioral entropy compared to the other groups. This is because ``Uncertainty ignorant'' entropies focus on exploring larger areas of uncertainty first as the information gain from almost/partially known area is significantly less than that from highly uncertain or unknown areas. 
Thereby, reducing uncertainty (Shannon's entropy) much faster than other groups. This is clearly evident under very noisy mapping ($\sigma^m =2$): it can be seen from Fig.~\ref{fig:results_poc_4} and Fig.~\ref{fig:results_poc_6}, that the ``Uncertainty ignorant'' group of Behavioral entropy clearly outperforms all other groups in all statistics,
resulting into a significantly lower range, mean and median iterations.  
This is because larger areas of uncertainty are explored first, not waiting for an area to get an almost/complete information gain before moving to the next. 
However, we observed that, for $\alpha \geq 10$, as the behavioral information gain of the explorer becomes $\approx 0$ for almost/partially known areas for the remaining frontiers, exploration terminates. We observed a completion (Shannon's entropy reduced) of around $75-90 \%$ for $\alpha = 10$. 
This behavior is well justified looking at the entropy for Bernoulli trials (Fig~\ref{fig:entropy}). As alpha increases, the tail ignorance grows and more likely outcomes also start getting ignored. 
Thus care needs to be taken to appropriately choose $\alpha$.

Although Renyi's entropy offers some tail ignorance with $\gamma > 1$, it is not enough to speed up exploration comparing to Behavioral entropy. This is evident from Fig.~\ref{fig:entropy}, where the variation for $\supscr{H}{R}$ from $1 \leftarrow \gamma$ (black $\supscr{H}{S}$ curve) to $\gamma \rightarrow \infty$ (green) is minimal as compared with Behavioral entropy $\supscr{H}{B}$ (red).

     

\color{black}
\section{Unity-ROS Experiments and results}
\label{sec:exp_results}

Here, we describe results in a ROS-Unity environment from the DARPA subterranean robotic challenge~\cite{darpasubt2021}.

\begin{figure}
    \centering
    \includegraphics[width=0.75\linewidth]{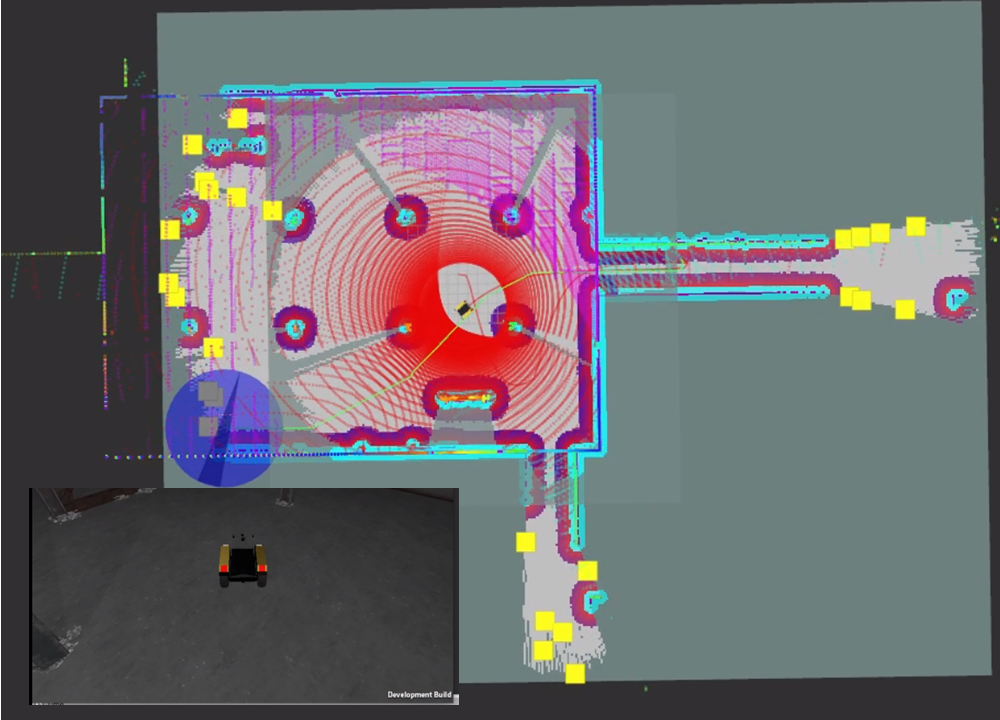}
    \caption{The unity-ROS environment used in DARPA subterranean challenge. Snapshot of execution shown with map and costmap, frontier clusters (yellow box), laser scans and robot going to region (blue circle), trying to follow a planned path (green line). }
    \label{fig:unity_execution}
    \vspace{-0.5cm}
\end{figure}

\paragraph*{Environment and Robot setup}
\label{sec:results_env_plan}
The simulated hardware experiments use a Clearpath Warthog robot\footnote{https://clearpathrobotics.com/warthog-unmanned-ground-robot/} fitted with Ouster LiDAR and IMU shown in Fig~\ref{fig:unity_execution} in ROS. 
{\color{black}Real hardware sensor and actuator models are used including noise sampled from real data. We test our exploration algorithm in the ``Urban Challenge'' Unity environment, part of the DARPA subterranean robotic challenge, see supplementary videos for scenarios that contains challenges like uneven surfaces, rubble and cluttered environment to make it as close to reality as possible.}
As we have an UGV that is incapable of climbing stairs, we will use a part of the environment whose ground truth is shown in Fig~\ref{fig:ground_truth}. The white areas are traversable whereas the black areas are obstacles or walls. 
We consider two different starting positions (green dots in Fig~\ref{fig:ground_truth})
for the robot facing towards left of the map and use Omnimapper~\cite{nieto2014IJRR} to map and localize the robot.


\paragraph*{Experiment setup}
\label{sec:exp_setup}
A snapshot of the execution is shown in Fig~\ref{fig:unity_execution}. 
The Omnimapper's updated occupancy map $\mathcal{M}$ and robot localization $x$ are shown in RViZ in Fig~\ref{fig:unity_execution} (left). 
Frontiers $f$ and perceived occupancy $\mathcal{W}$ are extracted from $\mathcal{M}$. 
The frontiers $f$ are clustered ($F$) (Yellow boxes in Fig~\ref{fig:unity_execution}) and Behavioral entropy $\supscr{H}{B}$ is calculated from $\mathcal{W}$. 
For each cluster in $F$, A Behavioral information gain $I^B$ is calculated as described earlier that corresponds to a beam based measurement model as in~\cite{ST-WB-DF:05}. 
Utilities $U$ are calculated from $I^B$ and estimated $x$ as in \eqref{eqn:utility_frontier}.
The explorer then picks a suitable goal region $B^x_r$ (shown as blue disk in Fig~\ref{fig:unity_execution}(right)) and sends it to the navigation manager, which in turn sends global and local plans $\eta$ to the controller. 
We use a Generalized Lazy Search (GLS)~\cite{mandalika2019gls} path planner for global planning, and MPPI~\cite{mppi_control:17} for local planning. 
The framework is visualised in~Fig~\ref{fig:framework}.

\begin{figure}
     \centering
     \begin{subfigure}[b]{0.23\textwidth}
         \centering
         \includegraphics[width=\textwidth]{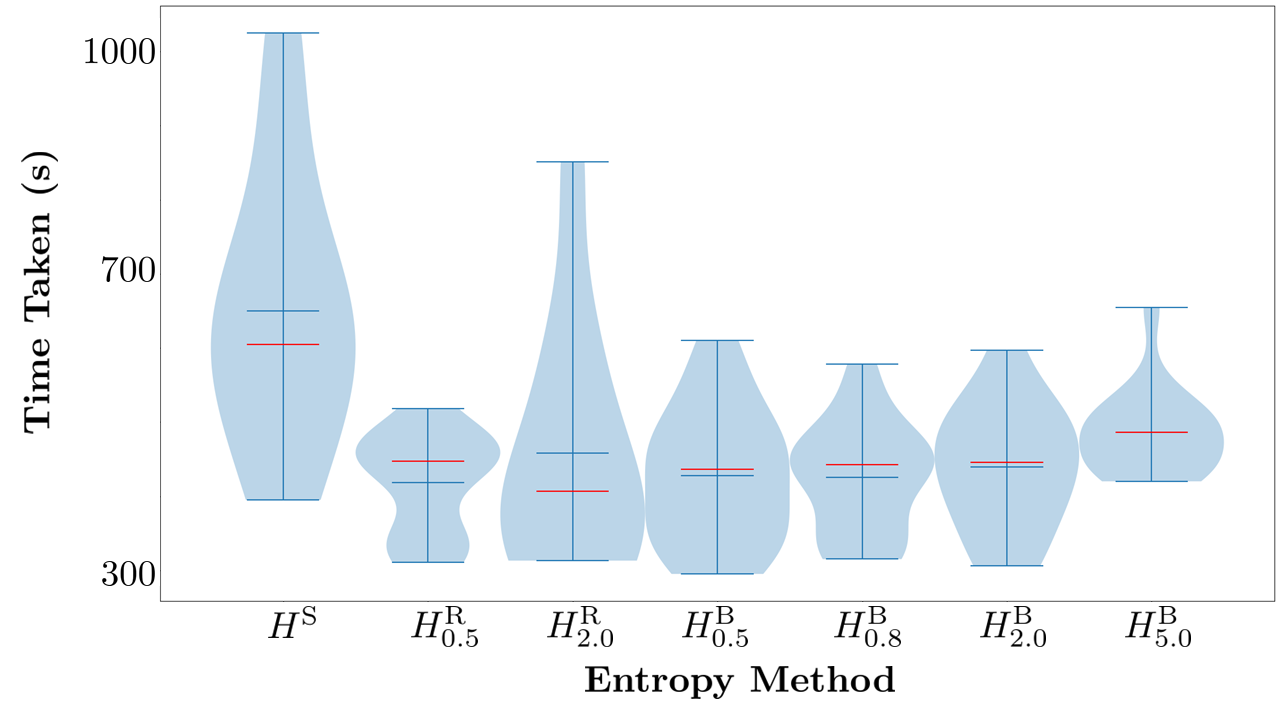}
         \caption{50 \% completion}
         \label{fig:results_uc1}
     \end{subfigure}
     ~
     \begin{subfigure}[b]{0.23\textwidth}
         \centering
         \includegraphics[width=\textwidth]{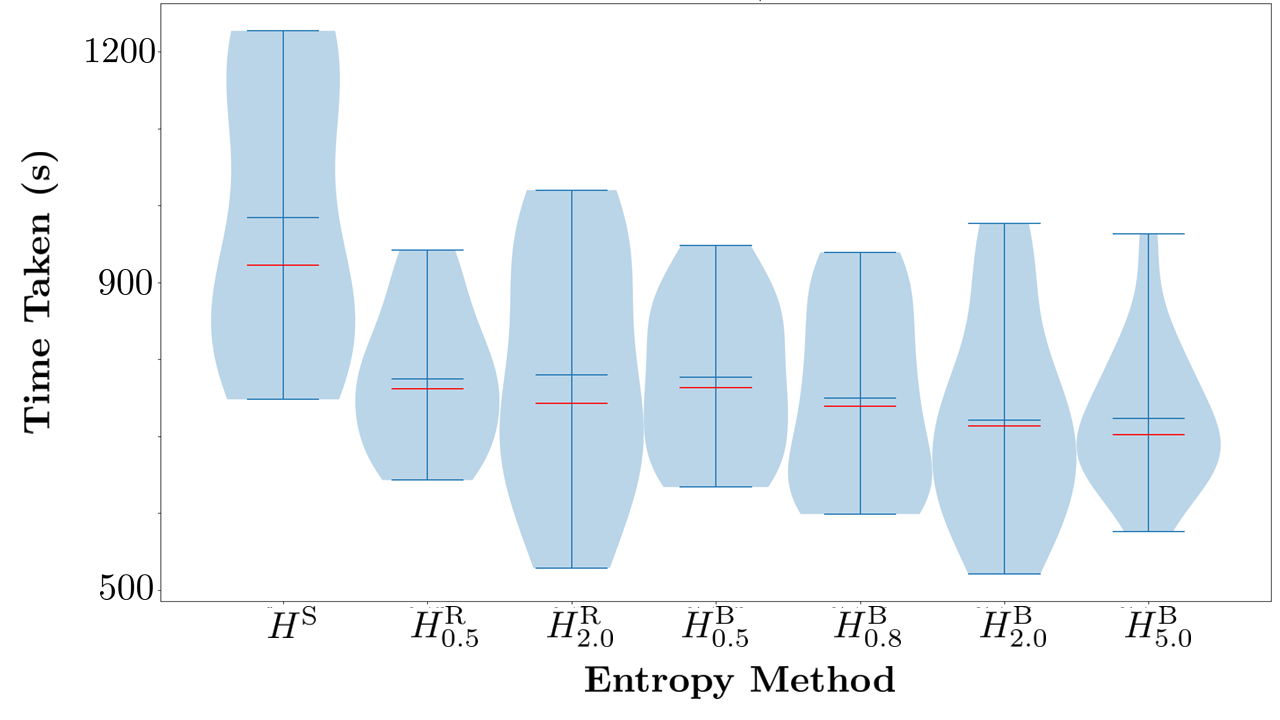}
         \caption{75 \% completion}
         \label{fig:results_uc2}
     \end{subfigure}
     
     \begin{subfigure}[b]{0.23\textwidth}
         \centering
         \includegraphics[width=\textwidth]{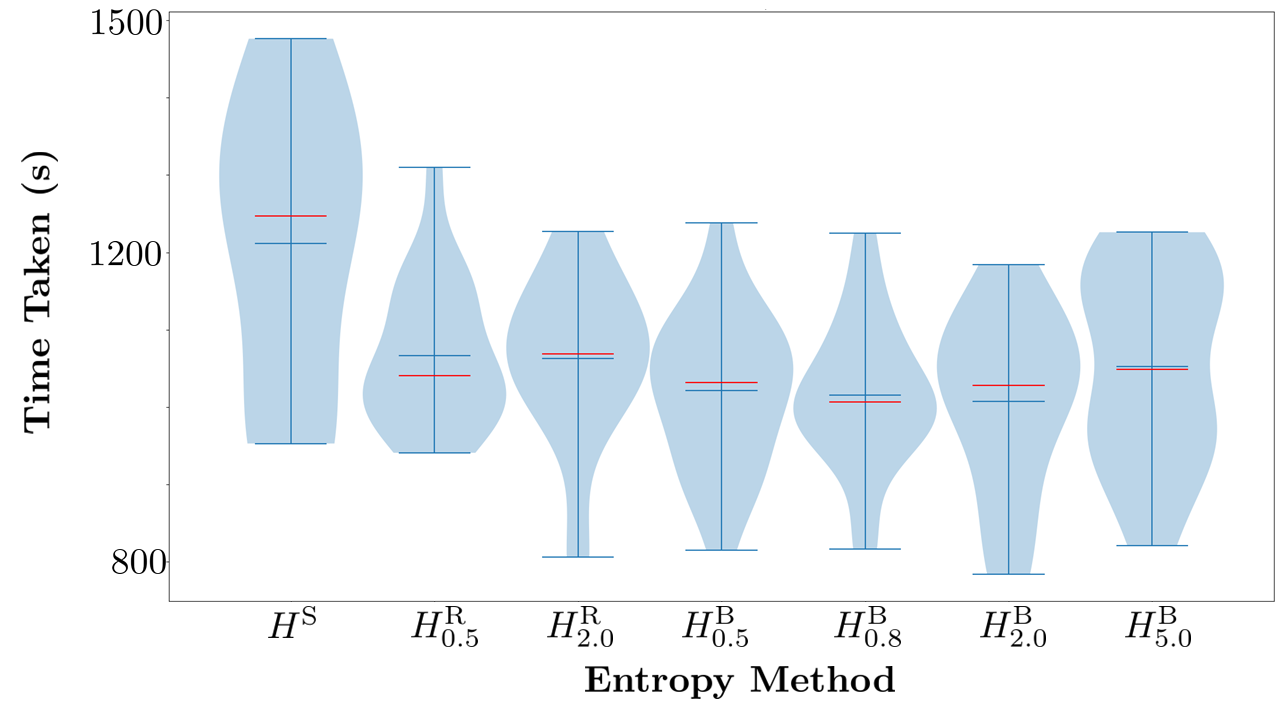}
         \caption{90 \% completion}
         \label{fig:results_uc3}
     \end{subfigure}
     ~
     \begin{subfigure}[b]{0.23\textwidth}
         \centering
         \includegraphics[width=\textwidth]{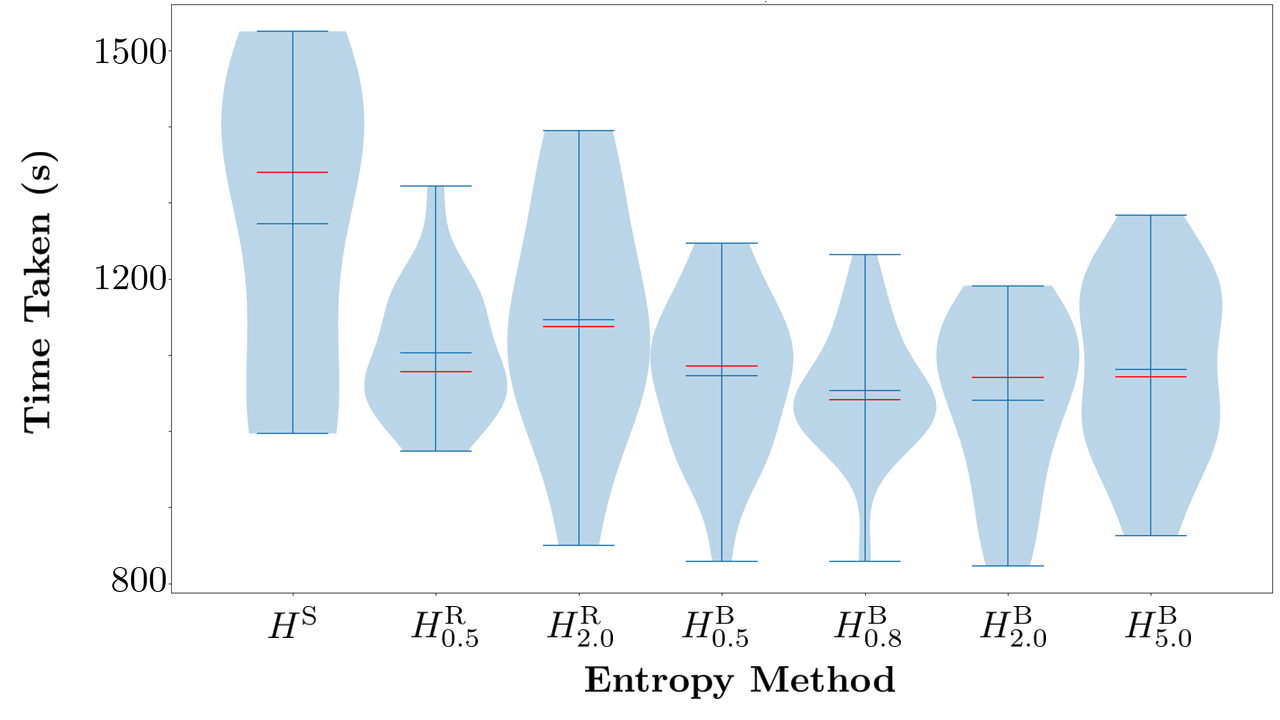}
         \caption{95 \% completion}
         \label{fig:results_uc4}
     \end{subfigure}
     \caption{Experimental results for the DARPA SubT environment.}
        \label{fig:results_uc}
        \vspace{-0.5cm}
\end{figure}

The explorer algorithm is implemented with a time limit of $30$ minutes, and terminates if there are no more frontiers or no perceived information to be gained. 
We consider $7$ different entropy parameter choices: Shannon's entropy $H^S$, Renyi's entropy $H^R$ with $\gamma \in \{0.5,2.0\}$, and Behavioral entropy $\supscr{H}{B}$ with $\alpha \in \{0.5,0.8,2.0,5.0\}$. We perform 10 trials at each starting position for each entropy or parameter choice and we measure the area explored and entropy reduced every $1$ second, and compute percentage completion from the ground truth. As the results for area explored and entropy reduced is similar, here we consider percentage entropy reduction as shown in Fig~\ref{fig:results_uc}. 
We again use violin plots to depict the distribution of time taken in seconds for $x \%,\, x \in \{50,75,90,95 \}$ completion. 
Due to inherent uncertainty in the mapping framework, it is very difficult to completely recover the ground truth for every cell of $\mathcal{M}$. Thus we go up to $95 \%$ completion in this case. 
As before, the $Y$-axis indicates the distribution of time in seconds for a particular group to complete a percentage of area/entropy. The interpretation of these plots is similar to those of Fig.~\ref{fig:results_poc}. 
\paragraph*{Results and Discussion}
\label{sec:results_uc}
We note that, unlike the POC experiments, here the inherent uncertainty and complexity of mapping, localization, planning, navigation and controls are present in the system.
Thus they add significant aleatoric uncertainty and make the trends in the results less apparent.
From the trials, the ``Uncertainty ignorant'' group ($\alpha,\gamma>1$) focuses on exploring larger areas of uncertainty first as the information gain from a partially known area is significantly less.  
This results in a ``Breadth-First-Search'' type of exploration. Whereas, the ``Uncertainty averse'' group ($\alpha,\gamma<1$) focuses on clearing an area before moving to the next one.
Thus, the information gain from a partially known area is still high and the utilities of frontiers in new far away areas are similar to the partially-known nearby frontiers. 
This group produced a ``Depth-First-Search'' type of exploration.  

The global and local planners performed better (time taken and path quality) for the goal regions that are closer to the current pose.  This results in larger wait times, poorer navigation and re-planning for further away goals, thus adversely affecting the ``Uncertainty ignorant'' group ($\alpha,\gamma>1$), which tend to produce further away goals than the other group.
Interestingly, Shannon's entropy is performing the worst among all the other choices as it suffers from implementation disadvantages of both the ``Uncertainty ignorant'' and ``Uncertainty averse'' cases: neither was an area explored completely in depth or breadth causing much longer execution times. Since Renyi's entropy has limited perceptiveness as compared to Behavioral entropy, we see that it is not performing as good as our proposed entropy, following a similar trend from POC experiments.
These behaviors are visualized in the accompanying video.
Thus, our proposed entropy outperforms the existing entropies used in the literature in both the POC environment and the ROS-Unity environment. Based on the results, we recommend $\supscr{H}{B}$ parameter $\alpha$ to be in the range of $0.5-0.8$ for good exploration conditions or about $2.0-3.0$ when there is excessive sensing or environmental noise. 

\section{Conclusions and Future Work}
This works presents a novel robotic exploration strategy, which relies on human models of perception uncertainty. To do this, we first define a novel human-inspired measure of uncertainty called ``Behavioral entropy'', which is theoretically analyzed and compared with other common entropy measures like Shannon's and Renyi's. We then used this family of entropies to define utility functions to guide exploration in a frontier-based approach. 
After this, we illustrate and show that our method is superior in proof of concept simulations as well as a DARPA Subterranean Challenge ROS-Unity simulation environment with a Clearpath Husky robot.
Future work will study the benefits of the spatial/temporal variation of $\alpha$ and conduct real world experiments. We would also like to investigate other applications of information-adaptive path planning algorithms using this entropy class. 
\bibliographystyle{IEEEtran}
 \bibliography{alias,main}
\end{document}